\documentclass[pmlr,twocolumn,10pt]{jmlr} 

\usepackage{footmisc}
\usepackage{colortbl}

\usepackage{subcaption}

\usepackage{booktabs}
\usepackage{siunitx}

\theorembodyfont{\upshape}
\theoremheaderfont{\scshape}
\theorempostheader{:}
\theoremsep{\newline}

\jmlrvolume{225}
\jmlryear{2023}
\jmlrsubmitted{LEAVE UNSET}
\jmlrpublished{LEAVE UNSET}
\jmlrworkshop{Machine Learning for Health (ML4H) 2023} 

\title[Multi-modal UMLS Graph Learning (MMUGL)]{Multi-modal Graph Learning over UMLS Knowledge Graphs}

\author{\Name{Manuel Burger} \Email{manuel.burger@inf.ethz.ch} \\
\Name{Gunnar Rätsch} \Email{raetsch@inf.ethz.ch} \\
\Name{Rita Kuznetsova} \Email{rita.kuznetsova@inf.ethz.ch} \\
\addr{Department of Computer Science, ETH Zürich, Switzerland}
}

\begin{document}

\maketitle

\begin{abstract}
Clinicians are increasingly looking towards machine learning to gain insights about patient progression. We propose a novel approach named \emph{Multi-Modal UMLS Graph Learning (MMUGL)} for learning meaningful representations of medical concepts using graph neural networks over knowledge graphs based on the unified medical language system. These concept representations are aggregated to represent a patient visit and then fed into a sequence model to perform predictions at the granularity of multiple hospital visits of a patient. We improve performance by incorporating prior medical knowledge and considering multiple modalities. We compare our method to existing architectures proposed to learn representations at different granularities on the MIMIC-III dataset and show that our approach outperforms these methods. The results demonstrate the significance of multi-modal medical concept representations based on prior medical knowledge.

We provide our code on GitHub\footnote{\url{https://github.com/ratschlab/mmugl}}.
\end{abstract}

\begin{keywords}
Knowledge Graphs, EHR, UMLS, Multi-Modal
\end{keywords} 
\section{Introduction}

Modern healthcare facilities record patient information as Electronic Health Records (EHR). EHR Datasets such as MIMIC-III~\citep{mimic-iii-johnson-nature16}, HiRID~\citep{hirid}, and eICU~\citep{eicu-pollard-nature18} enable modeling of disease progressions within a single hospital visit, for example in Intensive Care Units (ICU) \citep{mimic-benchmark-harutyunyan-nature19, yeche2021}, or progressions across multiple patient visits \citep{mime-choi-neurips18}. These progressions can be meaningfully encoded into patient representations using deep learning as shown by numerous prior works~\citep{retain-choi-neurips16, mime-choi-neurips18, gram-choi-kdd17, timeline-bai-kdd18, cgl-lu-ijcai21, chet-lu-aaai22}. This large body of work highlights the value of strong patient representations which aggregate information across entire patient histories from multiple hospital stays, enabling clinicians to model potential risks in various predictive tasks regarding patients' evolution. 

We see advantages of recent multi-modal approaches in the ICU setting~\citep{clinical-notes-ts-khadanga-emnlp19, husmann2022importance} and visit sequence modeling~\citep{cgl-lu-ijcai21}. In multi-modal EHR representation learning~\citep{graph-text-park-chil22}, we benefit from two modalities: structured EHR data (e.g., billing codes) and unstructured text information stored in rich clinical reports. Other modalities of medical data exist outside of in-hospital datasets, where a vast amount of prior medical knowledge is stored in static form in databases such as the Unified Medical Language System (UMLS)~\citep{umls-bodenreider-04}.

We identify two drawbacks of current UMLS-based approaches~\citep{cui2vec-beam-biocomp20, acronym-skreta-nature21, umls-embeddings-mao-amia20}. First, the approaches do not consider a complete set of relational information stored in UMLS (not considering multiple vocabularies) and solely use UMLS as a unified concept space. Second, prior solutions \citep{acronym-skreta-nature21, umls-embeddings-mao-amia20} specify the usage of hierarchical relations, which implies the use of an underlying graph in the form of a tree (single vocabulary). More complex graph structures inside and across vocabularies are thus omitted. 

We introduce \emph{Multi-Modal UMLS Graph Learning (MMUGL)} to overcome the previously stated limitations. \emph{MMUGL} is a novel approach for learning representations over medical concepts extracted from the UMLS Metathesaurus in the form of a complex knowledge graph and relations; extracted using a simple and ambitious procedure including a considerable set of vocabularies and all the relations across and within them.

We train a shared latent space~\citep{latent-kg-liu-neurips21} using auto-encoder pretraining techniques~\citep{gbert-shang-ijcai19} and we bridge the modality gap between structured EHR codes and unstructured text. The proposed approach includes rich prior knowledge, which is important in the medical domain. It deals with sample scarcity by relying on prior knowledge structure and pretraining techniques, and it leverages multiple modalities as inputs.

\subsection*{Contributions}

\begin{itemize}
   \item In Section~\ref{sec:concept-embedding}, we present a new medical knowledge representation technique using graph neural networks (GNN) on the UMLS Metathesaurus of unprecedented complexity. Previous research has explored this knowledge, but we go a step further and demonstrate that we can extract large and intricate knowledge graphs by taking into account a substantial portion of the entire UMLS Metathesaurus and incorporating a strong structural prior into our machine learning model, resulting in improved performance.

    \item We introduce a shared latent \emph{Concept Embedding}~(Sec.~\ref{sec:concept-embedding}) space and a shared \emph{Visit Encoder}~(Sec.~\ref{sec:visit-encoding}) to optimize the single latent space from any modality jointly in a parameter efficient manner. Opposed to the idea of separate model inputs per modality, we show the benefits of grounding all modalities with the same prior knowledge and training a single latent space for all input modalities (structured and unstructured EHR).

    \item In Section~\ref{sec:results} we demonstrate, that we strongly outperform prior graph-based works in pretraining and downstream tasks and can perform competitively with or outperform  prior work trained at a much larger scale of data. We show increased relevance of our approach at longer prediction horizons for readmission predictions.
    
\end{itemize}

\section{Related Work}
In the following, we introduce related work in EHR modeling, knowledge graph learning, and graph learning in the context of EHRs.

\paragraph{EHR} Various types of deep learning architectures have been proposed to learn representations at different granularities (patients, visits, histories, etc.) in EHR datasets. \citet{retain-choi-neurips16, timeline-bai-kdd18} propose EHR-specific visit sequence models. \citet{mime-choi-neurips18} propose to focus on the inherent structure of EHRs w.r.t. treatments, diagnosis, visits, and patients. \citet{med-bert-rasmy-nature21} adapt the masked language modeling approach to learn medical concept embeddings.

\paragraph{Multi-Modality}
Prior work has considered learning representations from either structured components of EHR data~\citep{mime-choi-neurips18, gram-choi-kdd17, med-bert-rasmy-nature21} or from unstructured clinical text reports~\citep{clinicalbert-alsentzer-acl19, bluebert-peng-bionlp19}. \citet{hcet-meng-biohi21, summarize-ehr-gong-pmlr18, clinical-intervention-suresh-pmlr17} have proposed multi-modal architectures and \citet{graph-text-park-chil22} go a step further and introduce even stronger structural priors, while considering the two modalities of structured EHR data, as well as unstructured clinical reports.

\paragraph{Knowledge Graphs and GNNs}
\label{sec:related-gnn}
A vast amount of static prior medical knowledge often remains untouched in current modeling approaches. This prior knowledge can be extracted and transformed into knowledge graphs~\citep{umls-kg-rotmensch-nature17, bert-clinical-kg-harnoune-science21}. Existing work in natural language processing has established the benefits of knowledge graph representations to various downstream applications~\citep{ernie-zhang-acl19, colake-sun-coling20, bert-mk-he-acl20, kepler-wang-tacl21}; where the most recent approaches include GNNs~\citep{qa-gnn-yasunaga-nacl21, dragon-yasunaga-neurips22}. We aim to leverage the success of GNNs to learn node (and edge) representations~\citep{gcn-kipf-iclr2017, graphsage-hamilton-neurips17, gat-velickovic-iclr18}. 

\paragraph{Graph Learning in EHR}
\emph{GRAM}~\citep{gram-choi-kdd17} proposed to include prior knowledge from medical ontologies such as the International Classification of Diseases (ICD).
To model structural and relational data explicitly, approaches have started to use GNNs. \citet{gbert-shang-ijcai19} proposed to use the Graph Attention~\citep{gat-velickovic-iclr18} operator together with an architecture to pretrain embeddings over two ontologies.

Other works learn over heterogeneous graphs with different types of nodes~\citep{cgl-lu-ijcai21, medgcn-mao-science22, hetero-gnn-med-recom-gong-science21}.
\citet{chet-lu-aaai22} construct a global graph of diseases, as well as dynamic local (within a single visit) subgraphs.
\citet{gct-choi-aaai20} focus on the EHR structure within a single visit.
Finally, \citet{sherbet-lu-ieee21, hyperbolic-icd-lu-bcb19} consider hyperbolic embeddings for medical ontologies. The learned embeddings can then be incorporated into task-specific architectures~\citep{med-bert-rasmy-nature21, gbert-shang-ijcai19, cgl-lu-ijcai21, kame-ma-cikm18} to improve outcome predictions in different healthcare settings.

Previous approaches do consider dataset-specific structures such as the hierarchical organization of EHRs (patients, visits, etc.) and co-occurrence information or structure coming from ontologies. However, the explored set of ontologies is usually kept small and most of them are tree-like structures. To the best of our knowledge, no prior work has considered using a GNN directly on top of a complex large-scale ontology such as the UMLS Metathesaurus and the complete set of unstructured relational information within it.

Further, while previous work considered multiple modalities, they use fusion approaches to join modalities, which tend to rely on larger architectures and thus require larger amounts of data to train effectively. Our work proposes to use the learned knowledge representations over the UMLS Metathesaurus as a single shared latent space for information coming from both the structured (billing codes) and unstructured modalities (clinical reports).

\section{Glossary}

We consider an EHR dataset of multiple patients and present the following terminology:

\begin{itemize}
    \item 
        \emph{Patient}: $p_i$ indexed by $i$
    \item 
        \emph{Visit}: a patient $p_i$ has one or multiple visits $v_{i,t}$ indexed by $t$. A visit contains a set of medical concepts $c \in \mathcal{C}_{i, t}$, the total set of medical concepts over the dataset is then $\mathcal{C} = \cup_{\forall i, t} \mathcal{C}_{i,t}$. A medical concept can be of different \emph{types} and we distinguish them by index $\mathcal{C}(*)$:

        \begin{itemize}
            \item
                \emph{Disease}: indexed by $d$ s.t. $\mathcal{C}_{i, t}(d)$ and $\mathcal{C}(d) = \cup_{\forall i, t} \mathcal{C}_{i,t}(d)$ the total set of disease concepts
           \item
                \emph{Medication}: (or prescriptions) with type $m$, similar to diseases we introduce $\mathcal{C}(m)$ and $\mathcal{C}_{i,t}(m)$
            \item
                \emph{Concept from clinical reports}: a set of medical concepts extracted from text data (clinical reports, Sec.~\ref{sec:nlp-pipeline}). The total set of considered medical concepts from text $\mathcal{C}(n) = \cup_{\forall i, t} \mathcal{C}_{i,t}(n)$ where the set $\mathcal{C}_{i,t}(n)$ is collected from all reports at a specific visit $t$ of patient $i$. The type is $n$ for text \emph{note}.
        \end{itemize}
        The vector representation of a visit considering data of a specific type $*$ is $\mathbf{v_{i,t}}(*) \in \mathbb{R}^k$.

    \item
        \emph{Ontology}: each ontology has a vocabulary $\mathcal{V}_{Ont}$ of medical concepts $c$ and defines some relation amongst the concepts of the vocabulary using an edge set $\mathcal{E}_{Ont}$, which constitutes the ontology graph $\mathcal{G}_{Ont} = (\mathcal{V}_{Ont},\, \mathcal{E}_{Ont})$. We consider the following ontologies/databases:
        \begin{itemize}
            \item $\mathcal{G}_{ICD}$ (International Classification of Diseases) where $\mathcal{C}(d) \subseteq \mathcal{V}_{ICD}$
            \item $\mathcal{G}_{ATC}$ (Anatomical Therapeutic Chemical Classification) where $\mathcal{C}(m) \subseteq \mathcal{V}_{ATC}$
            \item $\mathcal{G}_{UMLS}$ (Unified Medical Language System) where $\{\mathcal{C}(d) \cup \mathcal{C}(m) \cup \mathcal{C}(n)\} = \mathcal{C} \subseteq \mathcal{V}_{UMLS}$
        \end{itemize}

    \item \emph{Modality}: We leverage two EHR data modalities in this work: \emph{Structured} and \emph{unstructured} EHRs. We consider billing code information provided in a tabular fashion to be \emph{structured} EHR and clinical notes to be a form of \emph{unstructured} EHR.
\end{itemize}

\begin{figure*}[t]
  \centering
  \includegraphics[width=\linewidth]{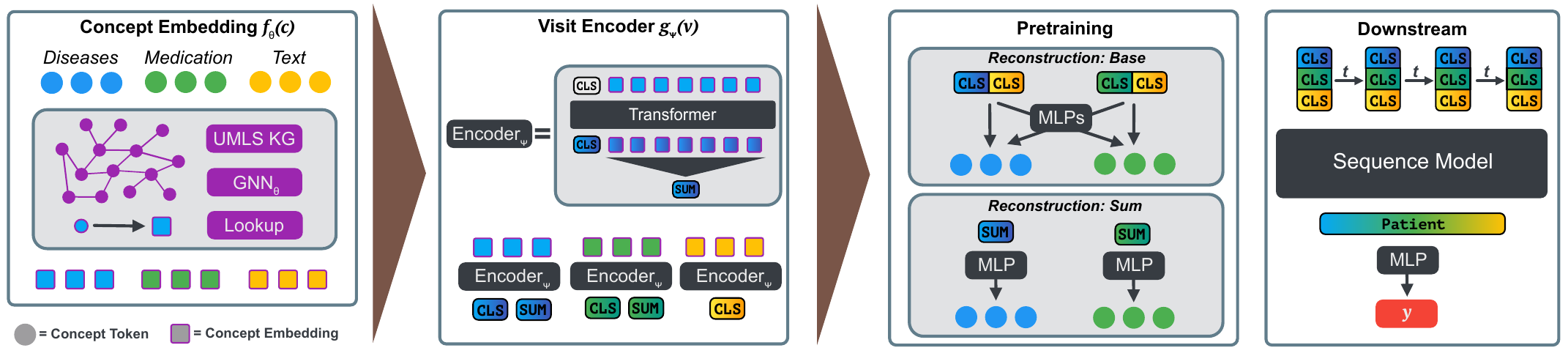}
  \caption{\emph{Model Architecture} brown arrows indicate information flow (Sec. \ref{sec:architecture}); visit information $v_{i,t}$ goes through the concept embedding $f_{\theta}$ (Sec.~\ref{sec:concept-embedding}), the representations get aggregated by $g_{\psi}$ (Sec.~\ref{sec:visit-encoding}) and then used by the pretraining single visit model and the downstream sequence models (Sec.~\ref{sec:predictors-training}). Data extraction illustrated in App.~\ref{apd:experimental-details}, Fig.~\ref{fig:data-extraction}.}
  \label{fig:architecture}
\end{figure*}

\section{Method}
\label{sec:architecture}
The architecture consists of three main components (\figureref{fig:architecture} provides an overview).

\begin{itemize}
    \item \textbf{Concept embedding module} $f_{\theta}(c): \mathcal{C}~\mapsto~\mathbb{R}^k$ (parametrized by $\theta$), which computes a representation for any given medical concept~$c$.
    \item \textbf{Visit encoding module} Assume $q = |\mathcal{C}_{i, t}|$ and $r \in \{2, 3\}$ the number of concept types considered (diseases and medications with or without concepts from text) then $g_{\psi}(v): \mathbb{R}^{q \times k} \mapsto \mathbb{R}^{r \times k}$ (parametrized by $\psi$), which, given all concept token representations of a single visit $v$ computes single representations for each different type of tokens thereof.
    \item \textbf{Predictor module} which performs either a pretraining task on a single visit or a downstream fine-tuning task across a sequence of visits. In either case, this module receives representations for each visit of a patient from the previous visit encoding module.
\end{itemize}

In the following subsections, we introduce the \emph{Concept Embedding} module (Sec.~\ref{sec:concept-embedding}), present how we extract richer concepts from clinical reports (Sec.~\ref{sec:nlp-pipeline}), encode the information (Sec.~\ref{sec:visit-encoding}), and perform predictions (Sec.~\ref{sec:predictors-training}).

\subsection{Concept Embeddings}
\label{sec:concept-embedding}

\label{sec:umls-kg}
We present our novel approach to rely on the UMLS Metathesaurus as a unified concept space to learn representations
for any general medical concept present in the database based on multiple modalities. Given that, we refer to our approach as \emph{Multi-Modal UMLS Graph Learning (MMUGL)}.

To constrain the number of concepts we consider from the database we use the set of clinical reports present in EHR datasets such as MIMIC-III~\citep{mimic-iii-johnson-nature16}. Using an extraction pipeline (Sec.~\ref{sec:nlp-pipeline}) we collect the set of medical concepts $\mathcal{C}(n)$.
The final vocabulary includes diseases and medications $\{\mathcal{C}(d) \cup \mathcal{C}(m) \cup \mathcal{C}(n)\} = \mathcal{C} = \mathcal{V}_{UMLS}$ and is used to construct $\mathcal{G}_{UMLS}$ by extracting all the edges in UMLS fully contained within the vocabulary. 

In UMLS many concepts are annotated with a short natural language description. We use \emph{SapBERT}~\citep{sapbert-liu-naacl21}, a fine-tuned language model to represent UMLS concepts, to initialize the node embeddings from these descriptions (ablation in Appendix \ref{apd:ablation-sapbert}). This contributes in two ways: (i) by not using trainable embeddings, we reduce the otherwise huge amount of free parameters given the large vocabulary $\mathcal{V}_{UMLS}$ (ii) we incorporate prior medical knowledge by considering the concept descriptions. We then train a multi-layer GNN on top of the extracted graph:

\begin{equation}
\label{eqn:umls-gnn}
    f_{\theta}(c) = \bigl( GNN_{\theta}(\mathcal{G}_{UMLS}) \bigr)[c]
\end{equation}

To retrieve a concept, we return its computed node embedding. We additionally found it to be beneficial for performance to consider two distinct stacks of GNN layers over the same graph and perform a \emph{Max-Pooling} operation after the final layer across the two stacks.

We consider two baseline graphs. First the tuple of the ICD and ATC hierarchy $ICD/ATC$ \citep{gbert-shang-ijcai19} and second an extension thereof including co-occurrence information $ICD/ATC-CO$ (e.g. \cite{cgl-lu-ijcai21}). More details are in Appendix \ref{apd:co-occurrence}. Additionally we consider a structureless baseline by replacing the concept embedding with an \emph{Embedding Matrix} (see App.~\ref{app:embedding-matrix}) which learns a representation for each concept without graph connectivity.

\subsection{Concept Extraction}
\label{sec:nlp-pipeline}

The goal of our approach is to include data from additional modalities such as the clinical reports found in EHR datasets. \emph{MMUGL} learns modality agnostic representations of medical concepts based on UMLS knowledge. It fuses discrete code information (e.g., ICD codes) with medical concepts extracted from text. The extraction with \emph{QuickUMLS}~\citep{quickumls-soldaini-sigir16} yields a \emph{set} of medical concepts $\mathcal{C}_{i,t}(n)$ based on the collection of clinical reports of that particular visit. Further, we perform a rule-based negation extraction using \emph{NegEx}~\citep{negex-chapman-science01}; for each concept, we extract a binary feature, whether it is negated or not, and concatenate it with its learned concept embedding (Eqn.~\ref{eqn:umls-gnn}). This is a crucial piece of information as clinical reports can both mention the existence or the absence of a certain condition.

\subsection{Visit Encoder}
\label{sec:visit-encoding}

The visit encoder $g_{\psi}(v)$ relies on the established Transformer architecture \citep{transformer-vaswani-neurips17}. Using a learned \texttt{CLS} token and without positional encoding, this serves as a set aggregation function. For each concept type in a given visit, we encode a separate representation using the same (weight-sharing) $\textit{Transformer}_{\psi}$ with parameters $\psi$ where $* \in \{d, m, n\}$. The concatenated representations $\mathbf{v_{i,t}}(*)$ encode the visit:

\begin{align}
    \label{eqn:type-visit-encoding}
    \mathbf{v_{i,t}}(*) &= \textit{Transformer}_{\psi}\Bigl( f_{\theta}\bigl( \mathcal{C}(*)_{i,t} \bigr) \Bigr) [\texttt{CLS}], \\
    \label{eqn:aux-visit-encoding}
    g_{\psi}(v) &= \left( \mathbf{v_{i,t}}(d), \mathbf{v_{i,t}}(m), \mathbf{v_{i,t}}(n) \right)
\end{align}

we can omit the multi-modal information from clinical reports (Eqn.~\ref{eqn:visit-encoding}), e.g., in cases where we
use a simpler baseline graph such as \emph{ICD/ATC} (Appendix Eqn.~\ref{eqn:icd-atc-embedding}) or in \emph{MMUGL without $\mathcal{C}(n)$}.

\begin{equation}
\label{eqn:visit-encoding}
    g_{\psi}(v) = \left( \mathbf{v_{i,t}}(d), \mathbf{v_{i,t}}(m) \right)
\end{equation}

\subsection{Predictors and Training}
\label{sec:predictors-training}

In the following, we introduce the pretraining module and downstream fine-tuning modules.

\subsubsection{Pretraining Module}\label{sec:pretraining-module}
We extend the auto-encoding pretraining approach developed by~\citet{gbert-shang-ijcai19} with the reconstruction loss $\mathcal{L}_{recon}$ and perform four different predictions (from each of the two \emph{types}, disease and prescription, as a source to either as the label) using distinct Multi-layer Perceptrons ($\textit{MLP}_{\bullet \rightarrow *}$ predicting type $*$ from representations of type $\bullet$) and attach a binary cross-entropy loss $\mathcal{L}_{BCE}$ to model multi-label classification.

\begin{align}
\label{eqn:auto-encoder-loss}
    \mathcal{L}_{recon} &= \sum_{\bullet, * \in \{d, m\}} \mathcal{L}'(\bullet, *), \\ 
    \nonumber \mathcal{L}'(\bullet, *) &= \mathcal{L}_{BCE}\Bigl( \textit{MLP}_{\bullet \rightarrow *} \bigl( \mathbf{v}(\bullet) \bigr) ,\, \mathcal{\mathcal{C}(*)} \Bigr)
\end{align}

During pretraining we randomly mask and replace tokens at the input in Eqn.~\ref{eqn:type-visit-encoding} (inspired by masked language modeling~\citep{bert-devlin-naacl19}).

\paragraph{Weighted reconstruction pretraining}

We consider a weighted version of Eqn.~\ref{eqn:auto-encoder-loss}:

\begin{equation}
\label{eqn:auto-encoder-loss-weighted}
    \mathcal{L}_{recon} = \sum_{\bullet, * \in \{d, m\}} w_{\bullet, *} \, \mathcal{L}'(\bullet, *)
\end{equation}

Some downstream tasks focus on disease diagnosis. Thus, we consider a tailored \emph{disease-focused} pretraining approach. In this setting, we omit the predictions (and loss signal) to medications and only predict diseases from either the visits aggregated disease or medication representation.
Meaning we set $w_{\bullet, d} = 1 \wedge w_{\bullet, m} = 0$. 

The contributions to the performance of this adaption are presented in Section~\ref{sec:results} and Appendix~\ref{apd:recon-loss}.

\paragraph{Sum Aggregation Loss} Due to the strong imbalance in the distribution of diseases and medications, we explore additional loss components to prevent the attention mechanism from overfitting to the most common tokens. Instead of taking the \texttt{CLS} token representation we take the sum over all tokens excluding \texttt{CLS} and again decode this unbiased aggregate using an \emph{MLP} to predict the set of diseases or prescriptions ($\setminus$ for set difference):

\begin{align}
\label{eqn:sum-pretraining-loss}
        \mathcal{L}_{sum} &= \sum_{* \in \{d, m\}} \mathcal{L}'(t), \\
        \nonumber \mathcal{L}'(*) &= \mathcal{L}_{BCE} \Bigl(\textit{MLP}^{\,sum}_{* \rightarrow *} \bigl( \mathbf{v}^{sum}(*) \bigr) ,\, \mathcal{C}(*) \Bigr), \\
        \nonumber \mathbf{v}^{sum}(*) &= \sum \biggl( \textit{Transf.}_{\psi}\Bigl(f_{\theta} \bigl( \mathcal{C}(*) \bigr) \Bigr) \setminus \{ \texttt{CLS} \} \biggr)
\end{align}

the idea is to ensure a more unbiased aggregation while still allowing the tokens to interact and impute masked or missing information. With this approach, we can induce a more dispersed distribution in the attention mechanism (App.~\ref{apd:sum-loss}).

\paragraph{Concepts from clinical reports}
In our approach \emph{MMUGL} we consider additional medical concepts extracted from text (clinical reports) and we concatenate the aggregated representation of these concepts for the respective visit $\mathbf{v}(n)$ to each of the two modalities at the input to the predictor \emph{MLP}. For example in the case of $\mathcal{L}_{recon}$:

\begin{align}
\label{eqn:text-loss}
    \mathcal{L}_{recon} &= \sum_{\bullet, * \in \{d, m\}} \mathcal{L}'(\bullet, *), \\
    \nonumber \mathcal{L}'(\bullet, *) &= \mathcal{L}_{BCE}\Bigl( \textit{MLP}_{\bullet \rightarrow *} \bigl( \mathbf{v}(\bullet) \oplus \mathbf{v}(n) \bigr) ,\, \mathcal{C}(*) \Bigr)
\end{align}

The final loss for pretraining $\mathcal{L}_{pre}$ is a combination of $\mathcal{L}_{recon}$ (Eqn.~\ref{eqn:auto-encoder-loss}, \ref{eqn:auto-encoder-loss-weighted}, \ref{eqn:text-loss}) and $\mathcal{L}_{sum}$ (Eqn.~\ref{eqn:sum-pretraining-loss}):

\begin{equation}
\label{eqn:final-pre-loss}
    \mathcal{L}_{pre} = \mathcal{L}_{recon} + \lambda \mathcal{L}_{sum},
\end{equation}

where $\mathcal{L}_{sum}$ is configured as a regularizer with hyperparameter $\lambda$ (for which we provide an ablation in Sec.~\ref{apd:sum-loss}).

\subsubsection{Downstream Modules} This work's contribution lies in learning concept representations over a knowledge graph from multiple modalities. We consider two prior architectures to perform time-series modeling and leave them mostly unchanged. It is intentional, that we do not propose a novel downstream architecture, but aim to show performance improvements alone through learning more robust and meaningful medical knowledge graph representations and aggregations thereof.

For medication recommendation, we employ an \emph{average pooling} scheme over the history used by~\citet{gbert-shang-ijcai19}. For all other tasks, we use an \emph{RNN} based model proposed by~\citet{cgl-lu-ijcai21}. More details are in Appendix \ref{apd:downstream-models}.

\begin{table}[tb]
    \renewcommand{\arraystretch}{0.9}
    \small
    \centering
    \caption{
        Performance comparison on \emph{Medication Recommendation}
        comparing to G-BERT~\citep{gbert-shang-ijcai19}; best in bold, $w_{\bullet, d}$ refers to Eqn.~\ref{eqn:auto-encoder-loss-weighted}. $\mathcal{C}(n)$ refers to med. concepts from text.
        Details in Section~\ref{sec:exp-med-recom}.}
    \setlength{\tabcolsep}{4pt}
    \begin{tabular}{l l l}
    \toprule
    Method & \textit{AuPRC} & \textit{F1} \\
    \midrule

G-BERT & 69.60 & 61.52  \\
    \emph{MMUGL w/o $\mathcal{C}(n)$}            & \textbf{72.27$\pm$0.21} & 63.39$\pm$0.04 \\
    \emph{MMUGL}                     & 72.10$\pm$0.17 & \textbf{63.43$\pm$0.06} \\
    \emph{MMUGL}, $w_{\bullet, d}=0$ & 72.10$\pm$0.19 & 63.34$\pm$0.23 \\

    \bottomrule
    \end{tabular}
    \label{tab:baseline-comparison-med}
\end{table} 
\begin{table*}[tb]
    \renewcommand{\arraystretch}{0.9}
    \small
    \centering
    \caption{Performance comparison on disease-related tasks with structural priors of EHR. Each section refers to results on the split and target code sets provided by the respective reference publication. \emph{MMUGL} shows the performance of our approach using a UMLS knowledge graph and information from clinical reports; $w_{\bullet, m}$ refers to Eqn.~\ref{eqn:auto-encoder-loss-weighted}; best in bold. Details in Section~\ref{sec:exp-disease} and Appendix \ref{apd:baselines}.}
    \begin{tabular}{l l l l l l}
    \toprule
    \bf Method & \multicolumn{2}{c}{\textbf{Heart Failure}} & \multicolumn{2}{c}{\textbf{Diagnosis}} \\
      & \textit{AuROC} & \textit{F1} & \textit{w-F1 (infl.)} & \textit{R@20} & \textit{R@40} \\
    \midrule

\multicolumn{5}{l}{\textbf{Reference: CGL \cite{cgl-lu-ijcai21}}} \\
    \arrayrulecolor{lightgray}\midrule\arrayrulecolor{black}
    GRAM \cite{gram-choi-kdd17}  & 82.82$\pm$0.06 & 71.43$\pm$0.05 & 21.06$\pm$0.19 & 36.37$\pm$0.16 & 45.61$\pm$0.27 \\
    Timeline \cite{timeline-bai-kdd18} & 80.75$\pm$0.46 & 69.81$\pm$0.34 & 16.83$\pm$0.62 & 32.08$\pm$0.66 & 41.97$\pm$0.74 \\
    MedGCN \cite{medgcn-mao-science22} & 81.25$\pm$0.15 & 70.86$\pm$0.18 & 20.93$\pm$0.25 & 35.69$\pm$0.50 & 43.36$\pm$0.46 \\
    RETAIN \cite{retain-choi-neurips16} & 82.73$\pm$0.21 & 71.12$\pm$0.37 & 19.66$\pm$0.58 & 33.90$\pm$0.47 & 42.93$\pm$0.39 \\
    CGL \cite{cgl-lu-ijcai21}  & 85.66$\pm$0.19 & 72.68$\pm$0.22 & 22.97$\pm$0.19 & 38.19$\pm$0.16 & 48.26$\pm$0.15 \\
    MedPath \cite{medpath-ye-www21} & 82.90$\pm$0.46 & 67.32$\pm$0.57 & - & - & - \\
    GraphCare \cite{graphcare-jiang-arxiv23} & 84.34$\pm$0.05 & 74.15$\pm$0.56 & 14.80$\pm$0.15 & 29.97$\pm$0.22 & 41.31$\pm$0.22 \\

     \emph{MMUGL}  & 87.19$\pm$0.21 & 74.46$\pm$0.41 & 25.81$\pm$0.17 & 41.02$\pm$0.10 & 52.41$\pm$0.14 \\
     \emph{MMUGL}, $w_{\bullet, m}=0$ & \textbf{87.60$\pm$0.40} & \textbf{74.71$\pm$0.72} & \textbf{26.40$\pm$0.02} & \textbf{41.54$\pm$0.17} & \textbf{53.02$\pm$0.37} \\

\midrule
    \multicolumn{5}{l}{\textbf{Reference: Chet \cite{chet-lu-aaai22}}} \\
    \arrayrulecolor{lightgray}\midrule\arrayrulecolor{black}
    Chet \cite{chet-lu-aaai22} & 86.14$\pm$0.14 & 73.08$\pm$0.09 & 22.63$\pm$0.08 & 37.87$\pm$0.09 & - \\
    \emph{MMUGL} & \textbf{87.67$\pm$0.36} & \textbf{75.49$\pm$0.48} & \textbf{25.55$\pm$0.48} & \textbf{40.95$\pm$0.13} & 52.51$\pm$0.33 \\

\midrule
    \multicolumn{5}{l}{\textbf{Reference: Sherbet \cite{sherbet-lu-ieee21}}} \\
    \arrayrulecolor{lightgray}\midrule\arrayrulecolor{black}
    G-BERT \cite{gbert-shang-ijcai19} & 83.61$\pm$0.18 & 72.37$\pm$0.46 & 22.28$\pm$0.25 & 36.46$\pm$0.15 & - \\
    Sherbet \cite{sherbet-lu-ieee21} & 86.04$\pm$0.16 & 74.27$\pm$0.07 & 25.74$\pm$0.04 & 41.08$\pm$0.08  & - \\
    \emph{MMUGL} & \textbf{87.58$\pm$0.32} & \textbf{75.65$\pm$1.04} & 25.78$\pm$0.08 & 41.00$\pm$0.08 & 52.81$\pm$0.31 \\
    \emph{MMUGL}, $w_{\bullet, m}=0$ & 87.47$\pm$0.31 & 75.25$\pm$0.26 & \textbf{26.79$\pm$0.18} & \textbf{42.03$\pm$0.04} & 53.55$\pm$0.32 \\

    \bottomrule
    \end{tabular}
    \label{tab:baseline-comparison-diag}
\end{table*} 
\section{Experimental Setup}
\label{sec:experiments}

We perform our experiments on the MIMIC-III~\citep{mimic-iii-johnson-nature16} dataset (version 1.4). Medications are mapped to the ATC hierarchy using the approach shared by~\citet{gbert-shang-ijcai19}. For pretraining we consider the training splits of the respective baselines as well as unused patients with only a single visit and thus not suitable for fine-tuning sequence tasks. We consider five different downstream tasks all trained using (binary) cross-entropy (binary/multi-label/multi-class). The tables show test set performance with standard deviations over three seeded training runs and we highlight the best results in bold font. In Appendix~\ref{apd:data}, \ref{apd:architecture}, \ref{apd:hyperparameters}, and \ref{apd:evaluation} we share data, training, architecture, and task details.

\paragraph{Training Procedure}
The proposed pretraining is task-agnostic and the same for all downstream tasks. After pretraining, the parameters of the concept embedding (Sec.~\ref{sec:concept-embedding}, GNN over the knowledge graph) are not fine-tuned for specific tasks. This contributes to the efficiency of our approach, as we can achieve all of the proposed performance gains without the need to retrain the large knowledge graph GNN for specific tasks. Only task-specific architectural components and the visit encoder (Sec.~\ref{sec:visit-encoding}) are fine-tuned.

\paragraph{Medication Recommendation}
\label{sec:med-recom-task}
To compare to the work by~\citet{gbert-shang-ijcai19} (who have shown improvements over any previously published results on this task) we benchmark the medication recommendation task. We use their provided preprocessed patient data derived from MIMIC-III.
The multi-label prediction task was evaluated on a sample-averaged Area under the precision-recall curve \emph{AuPRC}, as well as sample-averaged macro \emph{F1} score.

\paragraph{Heart Failure}
\label{sec:heart-failure-task}
This task has been benchmarked in \emph{CGL}~\citep{cgl-lu-ijcai21}, \emph{Chet}~\citep{chet-lu-aaai22}, and \emph{Sherbet}~\citep{sherbet-lu-ieee21}; who have performed extensive benchmarking against prior work.
We run their provided preprocessing and extract the used target code sets, as well as the computed patient splits. The binary classification is evaluated using \emph{F1} score and area under the receiver-operator curve \emph{AuROC}.

\paragraph{Diagnosis}
\label{sec:diagnosis-task} 
Similar to the previous heart failure task we compare to the results of \emph{CGL}, \emph{Chet}, and \emph{Sherbet}. We extract the target code sets and patient splits by running the provided preprocessing in each of the repositories to ensure comparability.
We consider thresholded \emph{weighted F1} (\emph{w-F1}) score, and to be comparable to~\citet{cgl-lu-ijcai21} we consider their adapted computation of \emph{F1}. The variant is slightly inflated by considering the number of ground truth positive labels for each sample. This avoids the need to set a threshold, but leaks the number of ground-truth positives to the evaluation; we refer to it as \emph{w-F1 (infl.)}. We also report recall at top $k$ predictions (according to model confidence); referred to as \emph{R@k} (e.g. \emph{R@20}).

\paragraph{Readmission}
We define a readmission task at a horizon $h$ for a given patient history at time $t$ and a subsequent admission at time $t_{readm.}$ the target is defined as $y_{readm.@h} = (t_{readm.} - t) < h$. (Emergency) readmissions are clinically highly relevant as shown by nationwide deployments of such systems in e.g. Scotland \citep{sparra-liley-medrxiv21}. We evaluate the performance using \emph{AuROC}.

\paragraph{Length of Stay}
We perform a length of stay prediction (\emph{LoS}) using a multi-class approach with 10 categories (\cite{pyhealth-yang-kdd23}). 0 for stays under 1 day, 1-7 for stays with the respective length in number of days, 8 for stays from 1 to two weeks, and 9 for stays over 2 weeks). To evaluate we use \emph{weighted-AuROC}.

\begin{table*}[tb]
    \renewcommand{\arraystretch}{0.9}
    \small
    \centering
    \caption{Ablation of different concept embeddings (Sec.~\ref{sec:concept-embedding}) considering data splits and target code sets of CGL~\citep{cgl-lu-ijcai21}. $w_{\bullet, m}$ refers to Eqn.~\ref{eqn:auto-encoder-loss-weighted}. $\mathcal{C}(n)$ refers to med. concepts from text. Best in bold. Details in Sec.~\ref{sec:exp-concept-emb} and Appendix \ref{apd:baselines}.}
    
    \setlength{\tabcolsep}{3pt}
    \begin{tabular}{l l l l l l l l}
    \toprule
    \bf Method & \multicolumn{2}{c}{\textbf{Heart Failure}} & \multicolumn{2}{c}{\textbf{Diagnosis}} & \multicolumn{1}{c}{\textbf{Readm.@1y}} & \multicolumn{1}{c}{\textbf{LoS}} \\
     & \textit{AuROC} & \textit{F1} & \textit{w-F1 (infl.)} & \textit{R@20} & \multicolumn{1}{c}{\textit{AuROC}} & \multicolumn{1}{c}{\textit{w-AuROC}} \\
    \midrule

Embedding Mat.              & 87.02$\pm$0.49 & 74.26$\pm$1.00 & 24.87$\pm$0.24 & 40.17$\pm$0.12 & 70.27$\pm$1.55 & 78.12$\pm$0.52 \\
    Node2Vec \emph{w/o $\mathcal{C}(n)$}    & 86.07$\pm$0.21 & 73.38$\pm$0.20 & 24.69$\pm$0.30 & 40.10$\pm$0.23 & 66.20$\pm$3.62 & 78.77$\pm$0.33 \\
    Cui2Vec \emph{w/o $\mathcal{C}(n)$} & 86.15$\pm$1.66 & 73.20$\pm$1.62 & 25.03$\pm$0.09 & 40.57$\pm$0.34 & 70.59$\pm$0.57 & 78.30$\pm$0.37 \\

GNN ICD/ATC  & 87.03$\pm$0.10 & 74.07$\pm$0.18      & 24.75$\pm$0.29 & 40.15$\pm$0.09 & 71.59$\pm$0.23 & 78.80$\pm$0.36 \\
    GNN ICD/ATC-CO  & 87.05$\pm$0.15 & 74.16$\pm$0.04   & 24.59$\pm$0.41 & 39.95$\pm$0.22 & 71.64$\pm$0.97 & 78.90$\pm$0.28   \\
    
\emph{MMUGL w/o $\mathcal{C}(n)$}  & 86.27$\pm$0.18 & 73.36$\pm$0.70 & 25.01$\pm$0.46 & 40.42$\pm$0.43 & 71.59$\pm$0.31 & 78.88$\pm$0.52 \\
    
    \midrule

Node2Vec \emph{with $\mathcal{C}(n)$}   & 87.03$\pm$0.35 & 74.25$\pm$0.38 & 24.75$\pm$0.87 & 40.02$\pm$0.77 & 71.23$\pm$0.50 & 81.20$\pm$0.26 \\
    Cui2Vec \emph{with $\mathcal{C}(n)$} &  87.51$\pm$0.10 & \textbf{74.71$\pm$0.24} & 25.84$\pm$0.11 & 41.02$\pm$0.15 & 72.57$\pm$0.20 & 81.96$\pm$0.11 \\
    
    \emph{MMUGL}  & 87.19$\pm$0.21 & 74.46$\pm$0.41                  & 25.81$\pm$0.17 & 41.02$\pm$0.10 & \textbf{73.26$\pm$0.18} & \textbf{82.18$\pm$0.37} \\
    \emph{MMUGL}, $w_{\bullet, m}=0$ & \textbf{87.60$\pm$0.40} & \textbf{74.71$\pm$0.72}    & \textbf{26.40$\pm$0.02} & \textbf{41.54$\pm$0.17} & 72.27$\pm$0.60 & 81.12$\pm$0.45 \\

    \bottomrule
    \end{tabular}
    
    \label{tab:concept-embedding-ablation}
\end{table*} 
\section{Results and Discussion}
\label{sec:results}

\paragraph{Medication Recommendation}
\label{sec:exp-med-recom}

Our method and training approach can outperform the previously published state-of-the-art results by~\citet{gbert-shang-ijcai19} (see \tableref{tab:baseline-comparison-med}). We note that the multi-modal approach with medical concepts from clinical reports cannot provide improvements on this task and data split (patients have high variation w.r.t.~the richness of available clinical reports); also see Appendix~\ref{apd:clinical-reports-performance}.

\paragraph{Disease Tasks}
\label{sec:exp-disease}

In \tableref{tab:baseline-comparison-diag} we present benchmarking results on two disease-related tasks (Sec.~\ref{sec:heart-failure-task}). We train and evaluate our models on the patient splits and code sets extracted by considering three different prior work implementations, which have performed extensive benchmarking on previous state-of-the-art methods.

Overall we can conclude improved performance against approaches using pretraining schemes \citep{sherbet-lu-ieee21}, including text data \citep{cgl-lu-ijcai21}, hyperbolic embeddings \citep{sherbet-lu-ieee21}, ontology graphs \citep{gram-choi-kdd17, gbert-shang-ijcai19, medgcn-mao-science22}, temporally localized graphs \citep{chet-lu-aaai22}, personalized patient graphs \citep{medpath-ye-www21}, and language model knowledge graphs \citep{graphcare-jiang-arxiv23}.

\paragraph{Concept Embedding Ablation}
\label{sec:exp-concept-emb}

In \tableref{tab:concept-embedding-ablation} we show ablations over different types of concept embeddings (Sec.~\ref{sec:concept-embedding}). All methods are trained with our proposed pipeline and only the concept embedding module is replaced. Our \emph{MMUGL} strongly benefits from richer multi-modal information coming from clinical reports and thus outperforms prior work (the multi-modal approach can also increase robustness w.r.t.~missing and erroneous information, Appendix~\ref{apd:missing-information}). We can see further improvements by tailoring our pretraining towards the downstream task by using \emph{disease-focused pretraining} (Eqn.~\ref{eqn:auto-encoder-loss-weighted} with $w_{\bullet, m} = 0$).

We compare to two alternative approaches without GNNs for learning concept embeddings by replacing the \emph{Concept Embedding}~\ref{sec:concept-embedding} module and performing the same proposed training procedure. As presented by~\citet{med-concepts-lee-jamia21} we pretrain our concept embeddings using \emph{Node2Vec}~\citep{node2vec-grover-kdd16}. Secondly, we compare to \emph{Cui2Vec}~\citep{cui2vec-beam-biocomp20}. \emph{Cui2Vec} consists of medical concept embeddings pretrained on a large-scale corpus using a \emph{Word2Vec}~\citep{w2v-mikolov-arxiv} style objective function. We show, that using our graph on the scale of 100,000 nodes and around 30,000 patients for pretraining, we can compete with or outperform an approach that used training data on the order of 60 million patients, 20 million clinical notes, and 1.7 million biomedical journal articles.

Our approach uses richer information from clinical reports and a larger concept vocabulary without introducing new parameters. Further, our approach is grounded in prior knowledge and is more general than previous work based solely on individual ontologies.

\paragraph{Readmission}
\label{sec:readmission}

\begin{figure}[t]
    \centering
    \includegraphics[width=\linewidth]{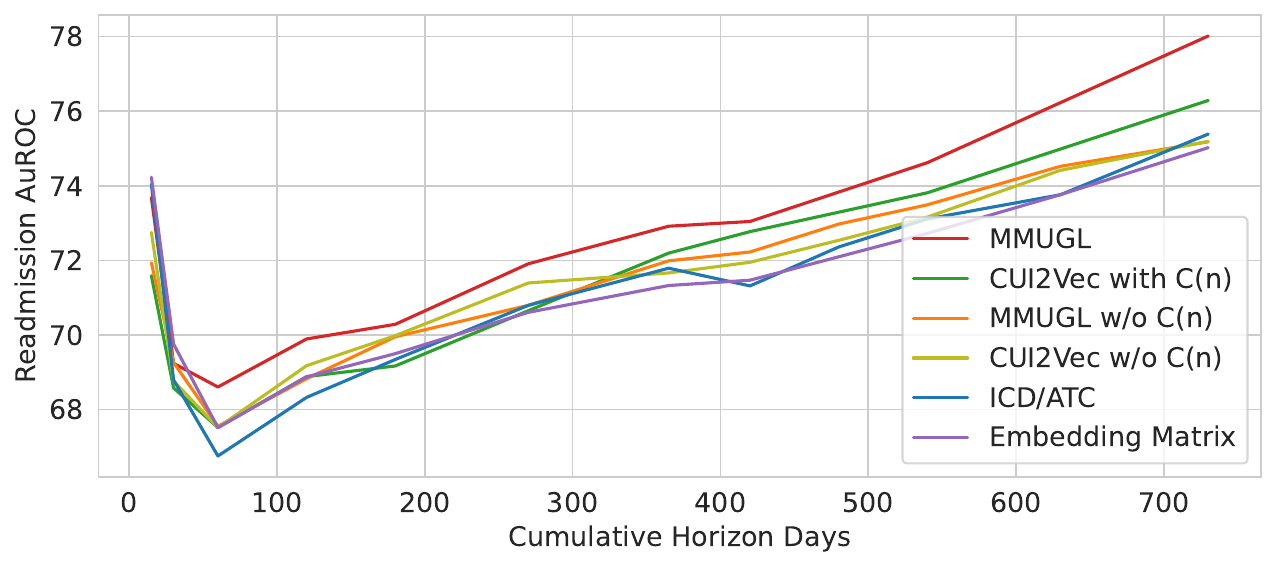}
    \caption{Readmission prediction at increasing cumulative horizons for different concept embeddings.}
    \label{fig:readmission-multiple-horizons}
\end{figure}

We benchmark readmission at different horizons and observe an increasing impact of our multi-modal knowledge graph at larger horizons (\figureref{fig:readmission-multiple-horizons}). \emph{MMUGL} combined graph-structured and multi-modal unified latent space is thus increasingly relevant for long-term risk predictions.

\tableref{tab:concept-embedding-ablation} shows performance at a relevant 1 year horizon \citep{sparra-liley-medrxiv21}. At a 15-day horizon we outperform very recent results by GraphCare~\citep{graphcare-jiang-arxiv23} using language model enhanced knowledge graphs who report $69.0$ \emph{AuROC} on MIMIC-III, where we achieve average \emph{AuROC} of $73.69$ with \emph{MMUGL} (and $71.93$ with \emph{MMUGL w/o $\mathcal{C}(n)$}). MedGTX by~\citet{graph-text-park-chil22} is a large fusion architecture and at a 30-day horizon they report $41.9$ \emph{AuPRC}, we outperform this result with our efficient shared latent space and achieve $45.27$ with \emph{MMUGL} (and $45.57$ with \emph{MMUGL w/o $\mathcal{C}(n)$}).

\paragraph{Summary}
\label{sec:discussion-summary}

\begin{table*}[tb]
    \renewcommand{\arraystretch}{0.9}
    \small
    \centering
    \caption{
        Performance contribution of different components: \emph{Base Emb. Matrix} uses an embedding matrix as \emph{concept embedding} (Sec.~\ref{sec:concept-embedding}, App.~\ref{app:embedding-matrix}) and is not pretrained. \emph{Pretr. Emb} pretrains the same embedding matrix (using Eqn.~\ref{eqn:final-pre-loss}). \emph{MMUGL} is our introduced knowledge graph based approach and $w_{\bullet, m}=0$ refers to disease-specific pretraining (Eqn.~\ref{eqn:auto-encoder-loss-weighted}). Best in bold, brackets indicate improvements over the baseline.
    }
    \setlength{\tabcolsep}{4pt}
    \begin{tabular}{l l l l l l}
    \toprule
    Component & \textbf{Heart Fail.} & \textbf{Diag.} & \textbf{Med. Rec.}  & \textbf{Readm.@1y} & \textbf{LoS} \\
              & \textit{AuROC} & \textit{w-F1 (infl.)} & \textit{AuPRC} & \emph{AuROC} & \emph{w-AuROC} \\
    \midrule

    Base Emb. Mat.      & 85.85 & 22.10 & 69.73 & 69.18 & 75.52 \\
    Pretr. Emb. Mat.    & 87.02 (+1.17) & 24.87 (+2.77) & 71.62 (+1.89)  & 70.27 (+1.09) & 78.12 (+2.60) \\
    \emph{MMUGL}        & 87.19 (+1.34)& 25.81 (+3.71) & \textbf{72.10 (+2.37)} & \textbf{73.26 (+4.08)} & \textbf{82.18 (+6.66)} \\
    \emph{MMUGL}, $w_{\bullet, m}=0$ & \textbf{87.60 (+1.75)} & \textbf{26.40 (+4.30)} & - & 72.27 (+3.09) & 81.12 (+5.60) \\

    \bottomrule
    \end{tabular}
    \label{tab:performance-ablation-components}
\end{table*} 
In Table~\ref{tab:performance-ablation-components} we summarize the performance benefits of different components for each benchmarked task. We can conclude that both pretraining (Sec.~\ref{sec:pretraining-module}), as well as our proposed knowledge graph concept embedding (Sec.~\ref{sec:concept-embedding}) improve performance across the set of considered tasks. We further note the relevance of disease-specific pretraining for tasks such as \emph{heart failure} or \emph{diagnosis}.

\paragraph{Interpretability}
\label{sec:interpretability-analysis-text-types}

Using the Transformer in our visit encoder (Sec. \ref{sec:visit-encoding}) over the shared latent space we can provide insights into how a trained model leverages the different modalities to perform predictions by analyzing attention scores. These scores can highlight to clinicians relevant medical concepts used for the predictions. In Appendices \ref{apd:single-patient-inter}, \ref{apd:text-concept-categories}, and \ref{apd:missing-information} we show examples of such an analysis.

\begin{figure}[t]
    \centering
    \includegraphics[width=\linewidth]{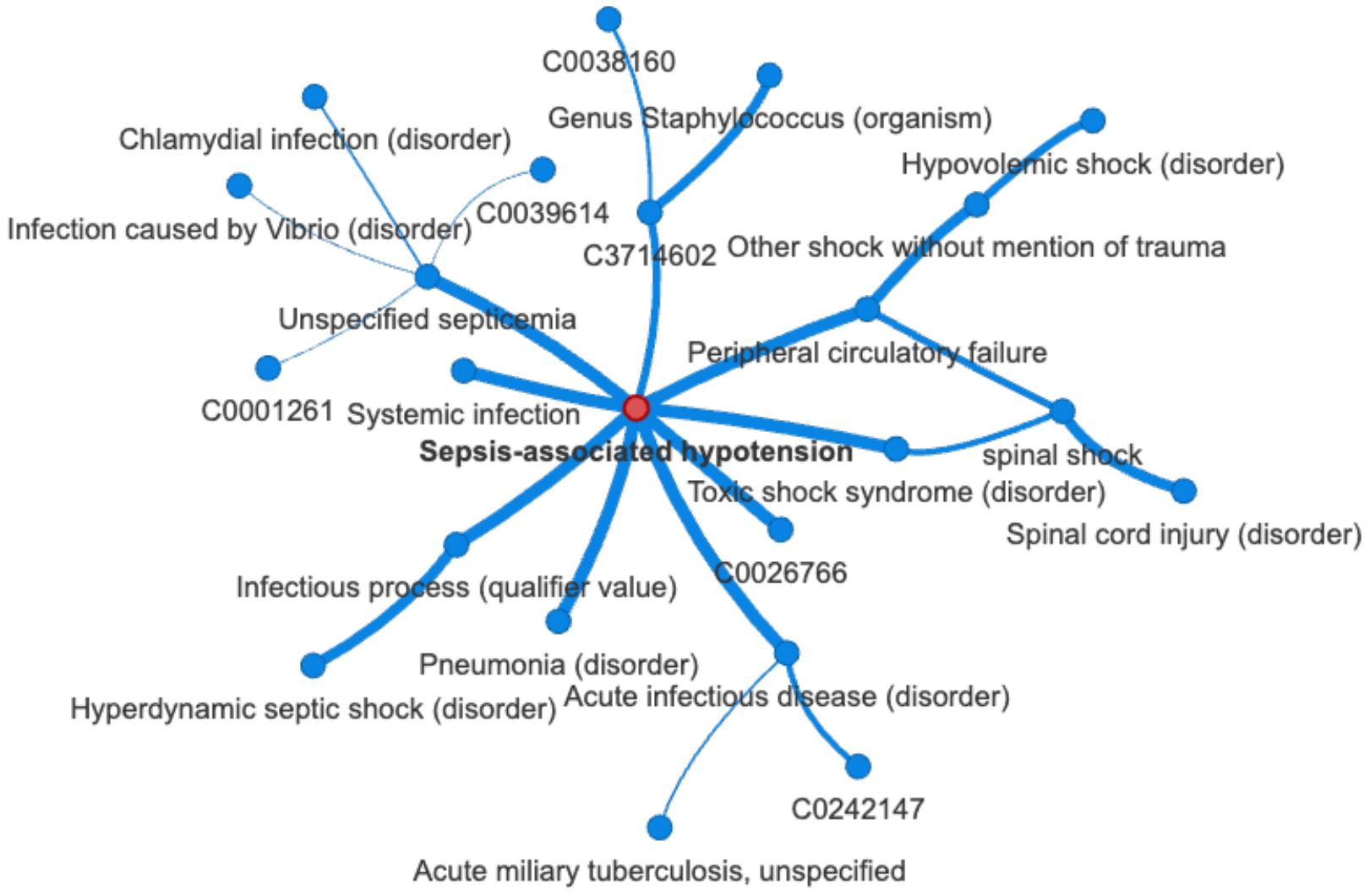}
    \caption{Subgraph relevant for the concept of \emph{Septic Shock} (\emph{Sepsis-associated hypotension}, UMLS \texttt{C0036983}) extracted using \emph{GNNExplainer}~\citep{gnnexplainer-ying-neurips19}. Edge thickness corresponds to the edge weight computed by \emph{GNNExplainer} and the explanation is a reduction of $9932$ nodes in a 3-hop neighborhood to $23$ related nodes.}
    \label{fig:septic-explanation-graph}
\end{figure}

\label{sec:graph-explain}
We can leverage the \emph{trained} knowledge graph to provide richer explanations to clinicians centered around important concepts. For example, the concept of \emph{Septic Shock} (UMLS \texttt{C0036983}) is connected to $9932$ nodes in a 3-hop neighborhood. Using \emph{ GNNExplainer}~\citep{gnnexplainer-ying-neurips19} we can retrieve a soft mask on the adjacency matrix (entries in $[0, 1]$), which represents an edge-weighted explanation subgraph. By thresholding this mask, we can retrieve an explicit subgraph explanation, where a user (e.g. a clinician) can control the size of the subgraph by adjusting the thresholding parameter. Figure~\ref{fig:septic-explanation-graph} shows an explanation graph around \emph{Septic Shock} with 23 nodes. The explanation subgraph aims to reduce the reconstruction error between the concept embedding $f_{\theta}(c)$ using the subgraph or the complete knowledge graph under the regularization constraints employed by \emph{GNNExplainer}.

\section{Conclusion}

We have introduced a novel parameter-efficient way to train a unified latent space for general medical knowledge from multiple modalities. We have demonstrated improved performance on downstream tasks by grounding our representations with prior knowledge from the UMLS Metathesaurus. Our extended pretraining approach and the corresponding results emphasize its importance in tackling the supervised label scarcity in the medical domain. The more generalized approach to medical concept representations can accommodate heterogeneous multi-modal EHR without complex fusion architectures. Future work could explore the integration of lab tests or procedure codes. Finally, more general medical concept embeddings can remove the need to perform mappings between different systems, as long as a map to UMLS exists.
   
Our results pave the way for future research to bridge the gap between within-visit modeling (e.g., 
ICU time-series models \citep{mimic-benchmark-harutyunyan-nature19}) and across-visit modeling, such as we benchmarked against in this work. Whereas disease and medication codes are usually assigned post-visit (for billing or archival purposes), many clinical reports are generated during the patient stays. To provide richer context information, future within-visit models might include patient histories and the multi-modal knowledge captured in our global concept representations.

\clearpage
\acks{
    This project was supported by grant \#2022-278 of the Strategic Focus Area “Personalized Health and Related Technologies (PHRT)” of the ETH Domain (Swiss Federal Institutes of Technology).
    
    Further, we would like to thank Hugo Yèche for his feedback during the revision process. Thanks go to Jonas Bokstaller and Severin Husmann whose theses have provided relevant insights.
}

\paragraph{Institutional Review Board (IRB)}
This research does not require IRB approval in the country in which it was performed.

\bibliography{burger23}

\clearpage
\appendix

\section{Experimental Details}
\label{apd:experimental-details}

\subsection{Dataset and Split details}
\label{apd:data}
A small overview of data and task statistics are provided in \tableref{tab:apd-data-statistics-disease}. Splits and target code sets have been extracted from the respective repositories\footnote{\url{https://github.com/jshang123/G-Bert},\\ \url{https://github.com/LuChang-CS/CGL},\\ \url{https://github.com/LuChang-CS/Chet},\\ \url{https://github.com/LuChang-CS/sherbet}}.

\begin{figure}[t]
    \centering
    \includegraphics[width=\linewidth]{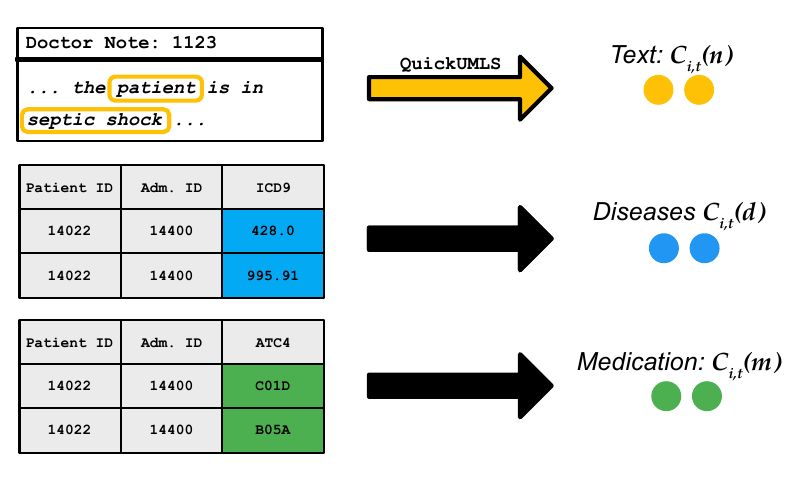}
    \caption{Illustration of the data extraction process (data is arbitrary). We extract structured EHR data in the form of billing codes from tables and use \texttt{QuickUMLS} to extract medical concepts from text (unstructured EHR) and match them against the UMLS database.}
    \label{fig:data-extraction}
\end{figure}

In Figure~\ref{fig:data-extraction} we provide a small illustration how we extract the different inputs to our model from the data.

\begin{table*}[h]
\floatconts
    {tab:apd-data-statistics-disease}
    {
        \caption{Dataset and task statistics for the disease-related tasks (Heart Failure Sec.~\ref{sec:heart-failure-task} and Diagnosis Sec.~\ref{sec:diagnosis-task}) and the medication recommendation task Sec.~\ref{sec:med-recom-task}. We compared to CGL by~\citet{cgl-lu-ijcai21}, Chet by~\citet{chet-lu-aaai22}, and Sherbet by~\citet{sherbet-lu-ieee21}, as well as G-BERT~\citep{gbert-shang-ijcai19} using the GitHub repository at \url{https://github.com/jshang123/G-Bert}. The splits and target code sets have been extracted by running the preprocessing code in their respective repositories; slight deviations from numbers reported in the referenced publications might be caused by a change of seed of the respective author or code modification post-publication which we can't reproduce. \emph{Pretraining} (Pre.) refers to the pretraining auto-encoder setup and \emph{Training} to the downstream medication recommendation task.}
    }
    {
        \begin{tabular}{l r r r r}
        \toprule
          & CGL & Chet & Sherbet & G-BERT \\
         \midrule
        \# Pretraining patients & 37'813 & 37'674 & 37674 & 34'965  \\
        \# Training patients & 6'155 & 6'009 & 6009 & 3'697  \\
        \# Validation patients & 125 & 493 & 493 & 1'058 \\
        \# Test patients & 1'221 & 1000 & 1000 & 1'059 \\
        \midrule
        \# Unique Diseases Pretrain & 6'984 & 6'984 & 6'984 & 1'997 \\
        \# Unique Diseases Training & 4'825 & 4'880 & 4'880 & 1'958 \\
        \midrule
        \# Unique Med. Pretrain & 348 & 348 & 348 & 468 \\
        \# Unique Med. Training & 348 & 348 & 348 & 145  \\
        \midrule
        \# UMLS Graph Nodes & 87'445 & 87'445 & 87'445 & 87'445 \\
        \bottomrule
        \end{tabular}
    }
\end{table*}

\subsection{Knowledge Graph Statistics}

The extracted knowledge graph contains 87'445 nodes, 261'212 edges with node degrees of $5.97\pm20.91$. The total vocabulary of all considered medical concepts is a subset of 21 \emph{UMLS Metathesaurus Vocabularies} (percentages in brackets, some concepts belong to multiple): SNOMEDCT\_US (46.75\%), ICD9\_CM (10.44\%), CCPSS (7.71\%), CSP (6.75\%), FMA (6.15\%), RXNORM (5.25\%), DXP (4.21\%), NCI\_CDISC (4.13\%), WHO (2.40\%), ATC (2.12\%), DRUGBANK (1.80\%), CPT, NOC, BI, CCS, ICNP, NIC, ICF, CCC, PCDS, RAM.

Given a patient split we compute the coverage over our vocabulary during pretraining and downstream training. Inclusion criterias causing differences between the two are availability of medication (req. for pretraining) and multiple visits (req. for downstream training).

\begin{itemize}
    \item \emph{All splits}: 91.59\% (pre), 71.24\% (down)
    \item \emph{Train}: 90.30\% (pre), 68.21\% (down)
    \item \emph{Validation}: 16.45\% (pre), 16.97\% (down)
    \item \emph{Test}: 37.68\% (pre), 38.47\% (down)
\end{itemize}

A percentage of concepts in the validation and test splits are unseen during training. Because of the graph structure, we can still learn meaningful representations for them:

\begin{itemize}
\item \emph{Validation}: 0.78\% (pre), 1.93\% (down)
\item \emph{Test}: 3.11\% (pre), 7.07\% (down)
\end{itemize}
 
\subsection{Architecture and Training}
\label{apd:architecture}

We perform early stopping based on the validation set loss both during pretraining and fine-tuning. The network is first fully pretrained until early stopped, the \emph{concept embedding} (Sec.~\ref{sec:concept-embedding}) backend is then frozen, the \emph{visit encoder} (Sec.~\ref{sec:visit-encoding}) is left trainable together with the downstream architecture to allow the attention mechanism to be fine-tuned to perform task-specific aggregations. We find a larger batch size (e.g. 32 or more) to be beneficial for better training stability.
Appendix~\ref{apd:hyperparameters} shows an overview of the hyperparameters, which have been tuned w.r.t.~validation set performance. 
\subsubsection{Downstream Models}
\label{apd:downstream-models}
We provide some details about the two downstream models used for sequence modeling based on prior work.

\paragraph{Average Pooling}\label{sec:avg-pooling-model} To compare to work by~\citet{gbert-shang-ijcai19} in medication recommendation, we consider their downstream architecture. Given a patient history of visits (of which we get the representations using modules from Sec.~\ref{sec:visit-encoding} and \ref{sec:concept-embedding}), we perform the same pooling scheme over the past and current visit to get a final representation which is used as input to an \emph{MLP} to perform a predictive task.

\paragraph{RNN}\label{sec:rnn-model} Based on the architecture by~\citet{cgl-lu-ijcai21} given a patient and sequence of past visits (obtained by encoding in Sec.~\ref{sec:visit-encoding}), we feed them through a \emph{GRU}~\citep{gru-cho-ssst14}. The hidden states at the output of the \emph{GRU} are aggregated using a temporal attention mechanism where the query is a trainable embedding. We perform a minor modification here w.r.t.~to the architecture by~\citet{cgl-lu-ijcai21} and introduce a hyperparameter $n_q$, which refers to the number of trainable queries. If more than one query is used, we aggregate the different temporal aggregations of each query to get a single representation of the entire past of the patient. This representation is used to perform a prediction into the future using a \emph{MLP}. 
\subsubsection{Baseline Graphs}
\label{apd:co-occurrence}
The following two paragraphs introduce the baseline graphs considered, namely a combination of the ICD (diseases) and ATC (treatments) hierarchies, and an extension with co-occurrence information.

\paragraph{ICD/ATC Hierarchies}
\label{sec:icd-atc-hierarchies}

Based on the work done by~\citet{gbert-shang-ijcai19}, we consider the two tree hierarchies ICD\footnote{We consider the 9th revision, as of working on MIMIC-III} for diseases and ATC for medications. In this case, we consider $c \in \{\mathcal{C}(d) \cup \mathcal{C}(m)\}$. We compute the node embeddings $\mathbf{N}_{*}$ ($\oplus$ for concatenation):

\begin{align}
    \label{eqn:icd-gnn} \mathbf{N}_{\mathcal{C}(d)} &= GNN_{\theta_1}(\mathcal{G}_{ICD}), \\
    \label{eqn:atc-gnn} \mathbf{N}_{\mathcal{C}(m)} &= GNN_{\theta_2}(\mathcal{G}_{ATC}), \\
    \label{eqn:icd-atc-embedding} f_{\theta}(c) &= Lookup(\mathbf{N}_{\mathcal{C}(d)} \oplus \mathbf{N}_{\mathcal{C}(m)})(c)
\end{align}

where we use a distinct (parametrized by $\theta_1$ and $\theta_2$) multi-layer GNN for each of the two hierarchies (Eqns.~\ref{eqn:icd-gnn},~\ref{eqn:atc-gnn}) and then perform a \emph{lookup} (retrieve nodes by index)
against the resulting node embeddings (Eqn.~\ref{eqn:icd-atc-embedding}). In this case, we initialize all of the nodes with randomly initialized trainable embeddings. We refer to this approach to learn concept embeddings with \emph{ICD/ATC}.

We can additionally consider co-occurrence information (e.g., \cite{cgl-lu-ijcai21, rrr-liu-neurips22}) to connect the two hierarchies. We refer to this approach with \emph{ICD/ATC-CO} and provide details in the following paragraph.

\paragraph{ICD/ATC with Co-Occurrence}
Similar to work done by~\citet{cgl-lu-ijcai21} or~\citet{rrr-liu-neurips22} we can
additionally consider co-occurrence information present in our dataset. We construct a new graph $\mathcal{G}_{ICD/ATC-CO}$ which contains multiple sets of nodes and edges. The node sets are the ICD and ATC tree hierarchy nodes, while the edge sets consist of
the two ontologies and four co-occurrence edge sets; one for co-occurrence within each of the two ontologies and one (directed) from
each of the two to the other. We then compute a heterogeneous (nodes of different types) multi-layer GNN (see~\cite{hetero-gcn-schlichtkrull-springer18, medgcn-mao-science22}) over these node and edge sets, where each edge set is associated with its own parametrized graph convolution operator. As a result, we compute multiple different embeddings for a given node in each layer, which are summed. Co-Occurrence edges can additionally be weighted by computing a count over the dataset (training split) and normalizing s.t. incoming edges sum to one. Such weights can be considered by the GNN by multiplying messages from neighboring nodes with the corresponding weight. Again we have $c \in \{\mathcal{C}(d) \cup \mathcal{C}(m)\}$:

\begin{equation}
\label{eqn:co-gnn}
    f_{\theta}(c) = GNN_{hetero}(\mathcal{G}_{ICD/ATC-CO})(c)
\end{equation} 
\subsubsection{GNN Architecture}
\label{apd:gnn-architecture}

In Section \ref{sec:concept-embedding} we use a parametrized GNN  in Eqns.~\ref{eqn:icd-gnn}, \ref{eqn:atc-gnn}, and~\ref{eqn:umls-gnn}. We use Pytorch Geometric \citep{pyg-fey-iclr19} to implement these networks and based on our hyperparameter searches in Appendix \ref{apd:hyperparameters} we settled on using the graph convolution operator GraphSAGE as introduced by \citet{graphsage-hamilton-neurips17}.

The ICD and ATC hierarchical ontologies or our complex UMLS based knowledge graph are passed to the GNN considering all edges as undirected. In the case of multiple GNN layers we use a non-linear ReLU activation after all but the last layer. The representations for each medical concept of an ontology or the knowledge graph at the GNN output are cached and used to retrieve concept embeddings for further processing by the Visit Encoder (Sec. \ref{sec:visit-encoding}) module. 
\subsection{Hyperparameters}
\label{apd:hyperparameters}
In \tableref{tab:apd-hp-pretraining,tab:apd-hp-medication-recommend,tab:apd-hp-heart-failure,tab:apd-hp-diagnosis} we present an overview of the model hyperparameters. Final choices based on validation set performances have been marked in bold font.

\paragraph{Hardware} A typical training is finished in under a day. Depending on the task and set of considered input modalities it can be much faster. We trained our models using mostly \texttt{Nvidia RTX2080Ti} GPUs with 11GB of dedicated GPU memory; some larger models, which included medical concepts extracted from text have been trained on \texttt{Nvidia Titan RTX} GPUs with 24GB of dedicated GPU memory. We use 2-6 worker processes and around 32-64GB of main memory.

\begin{table*}[p]
\floatconts
    {tab:apd-hp-pretraining}
    {
        \caption{Tested hyperparameters to optimize pretraining performance. Values that we have settled on (due to performance or training feasibility) are marked in bold. Graph depth of 0 means we have used a simple embedding layer (i.e. lookup table with trainable embeddings) instead of running a GNN over the node embeddings.}
    }
    {
        \begin{tabular}{l c}
        \toprule
        \textbf{Parameter} & \textbf{Values} \\\midrule
        Effective Batch Size & (1, 2, 4, 8, 16, \textbf{32}, \textbf{64}, 128) \\
        Hidden Dim. & (32, 64, 128, 196, \textbf{256}) \\
        GNN Depth & (0, 1, \textbf{2}, \textbf{3}, 4, 5, 6) \\
        GNN Activation Function & ReLU \\
        Graph Convolution & (GCN, GAT, GINConv, \textbf{GraphSAGE}) \\
        Graph Staged (like \citet{gbert-shang-ijcai19}) & (yes, \textbf{no}) \\
        Graph Stacks & (\textbf{1}, \textbf{2}, 4); 2 for UMLS graphs \\
        Transformer \# Att. Heads & (1, \textbf{2}, 4) \\
        Transformer \# Layers & (\textbf{1}, 2) \\
        Learning Rate (Adam \citep{adam-kingma-and-ba-iclr15-poster}) & (1e-4, \textbf{5e-4}, 7.5e-4) \\
        Visit Decoder Hidden & (\textbf{128}, 256) \\ 
        Sum Loss $\lambda_{Pool}$ & [0.0, 100.0]: \textbf{0.25} \\
Co-Occurrence Links (weighted) & (\textbf{yes}, no) \\
\bottomrule
        \end{tabular}
    }
\end{table*}

\begin{table*}[p]
\floatconts
    {tab:apd-hp-medication-recommend}
    {
        \caption{Tested hyperparameters to optimize the medication recommendation task (Sec. \ref{sec:med-recom-task}). If a relevant parameter is not mentioned explicitly here, we have defaulted to the choices presented in \tableref{tab:apd-hp-pretraining}. Graph depth of 0 means we have used a simple embedding layer instead of running a GNN over the node embeddings.}
    }
    {
        \begin{tabular}{l c}
        \toprule
        \textbf{Parameter} & \textbf{Values} \\\midrule
        Effective Batch Size & (1, 2, 4, 8, 16, \textbf{32}, 64) \\
        Freeze Graph & (\textbf{yes}, no) \\
        Freeze Encoder & (yes, \textbf{no}) \\
        Graph Depth & (0, 1, \textbf{2}, 3, 4, 5, 6) \\
        Learning Rate (Adam \citep{adam-kingma-and-ba-iclr15-poster}) & (1e-3, 5e-4, \textbf{1e-4}) \\
        \bottomrule
        \end{tabular}
    }
\end{table*}

\begin{table*}[p]
\floatconts
    {tab:apd-hp-heart-failure}
    {
        \caption{Tested hyperparameters to optimize the heart failure task (Sec. \ref{sec:heart-failure-task}). If a relevant parameter is not mentioned explicitly here, we have defaulted to the choices presented in \tableref{tab:apd-hp-pretraining}.}
    }
    {
        \begin{tabular}{l c}
        \toprule
        \textbf{Parameter} & \textbf{Values} \\\midrule
        Effective Batch Size & \textbf{32} \\
        Freeze Graph & (\textbf{yes}, no) \\
        Freeze Encoder & (yes, \textbf{no}) \\
        Graph Depth & \textbf{2} \\
        Learning Rate (Adam \citep{adam-kingma-and-ba-iclr15-poster}) & (\textbf{1e-5}, 2e-5, 5e-5, 1e-4, 5e-4) \\
        \# Temporal query vectors $n_q$ & (\textbf{1}, 2, 4) \\
        Classification MLP Hidden Layers & \textbf{2x128} \\
        \bottomrule
        \end{tabular}
    }
\end{table*}

\begin{table*}[p]
\floatconts
    {tab:apd-hp-diagnosis}
    {
        \caption{Tested hyperparameters to optimize the diagnosis prediction task (Sec. \ref{sec:diagnosis-task}). If a relevant parameter is not mentioned explicitly here, we have defaulted to the choices presented in \tableref{tab:apd-hp-pretraining}. 0 hidden layers in the classification MLP essentially means there is a single linear classification layer, so no non-linear activations are involved and there are no hidden layers.}
    }
    {
        \begin{tabular}{l c}
        \toprule
        \textbf{Parameter} & \textbf{Values} \\\midrule
        Effective Batch Size & \textbf{32} \\
        Freeze Graph & (\textbf{yes}, no) \\
Freeze Encoder & (yes, \textbf{no}) \\
        Graph Depth & (2, \textbf{3}) \\
        Learning Rate (Adam \citep{adam-kingma-and-ba-iclr15-poster}) & (1e-5, 2e-5, 5e-5, \textbf{1e-4}, 5e-4) \\
        \# Temporal query vectors $n_q$ & (1, 2, \textbf{4}) \\
        Classification MLP \# Hidden Layers & (\textbf{0}, 1, 2) \\
        Classification MLP Hidden Layer Dim. & (64, 128, 256, 512) \\
        \bottomrule
        \end{tabular}
    }
\end{table*} 
\subsection{Tasks and Evaluation}
\label{apd:evaluation}

In the following, we provide a more detailed overview of the benchmarked downstream tasks (Sec. \ref{sec:experiments}) and the evaluation thereof.

\subsubsection{Medication Recommendation}
We benchmark the medication recommendation task based on preprocessed data by \citet{gbert-shang-ijcai19}. The task is to predict a set of medications (ATC level 4 codes) given a patient's history and the current diagnosis (assigned ICD codes). Given a patient $i$ and a trained predictor $\hat{h}$ we can formalize as follows:

\begin{align}
\label{eqn:detail-med-recommend}
    \hat{\mathcal{C}}_{i, t}(m) &= \hat{h} \bigl( \mathcal{C}_{i, 0\ldots t-1}(*),\, \mathcal{C}_{i, t}(d) \bigr) \\
    & \text{where}\,* \in \{{d, m, n}\} \nonumber
\end{align}

Given that this is a multi-label prediction we consider sample-averaged scores. Due to a significant imbalance in the distribution of medication codes, we use the \emph{F1} score for thresholded hard predictions and the area under the precision-recall curve (\emph{AuPRC}) for unthresholded confidence scores. This is in line with the evaluation by \citet{gbert-shang-ijcai19}.

\subsubsection{Heart Failure}
This is a binary prediction task as already benchmarked by many prior works on the MIMIC-III \citep{mimic-iii-johnson-nature16} dataset. The task is to predict the risk of heart failure for a patient in a future visit given the patient's history. The label is extracted from the set of assigned ICD codes by matching with the prefix $428$ after stripping the codes of any special characters. Let $y_{i, t}$ be the target label and it is $1$ if there exists a code $c \in \mathcal{C}_{i, t}(d)$ which has the prefix $428$. For a patient $i$ and trained predictor $\hat{h}$ we can formalize as follows:

\begin{align}
\label{eqn:detail-heart-failure}
    \hat{y}_{i, t} &= \hat{h} \bigl(  \mathcal{C}_{i, 0\ldots t-1}(*) \bigr) \\
    & \text{where}\,* \in \{{d, m, n}\} \nonumber
\end{align}

The task with mild label imbalance is evaluated using \emph{F1} score and area under the receiver-operator curve (\emph{AuROC}) for untresholded performance evaluation; this is in line with work by \citet{cgl-lu-ijcai21, sherbet-lu-ieee21, chet-lu-aaai22, retain-choi-neurips16} and others.

\subsubsection{Diagnosis}
This is a multi-label prediction over a set of diseases. Given a patient's history we predict the set of potential diseases for an upcoming visit . For a patient $i$ and trained predictor $\hat{h}$ we can formalize as follows:

\begin{align}
\label{eqn:detail-diagnosis-task}
    \hat{\mathcal{C}}_{i, t}(d) &= \hat{h} \bigl( \mathcal{C}_{i, 0\ldots t-1}(*) \bigr) \\
    & \text{where}\,* \in \{{d, m, n}\} \nonumber
\end{align}

This task might not seem very sensible at first as we cannot expect to reliably predict accidents that cause a hospital visit based on past EHR records. However, this is useful to catch chronic diseases and re-occurring patient patterns. Such a model's predictions could serve as a high-level aggregation of all EHR records for a specific patient. A doctor can get a very quick assessment of the potential risks for a patient upon admission and can tailor further investigations to this.

Due to the extreme imbalance over the very large set of potential labels we use \emph{weighted-F1} score. To assess the unthresholded model confidence scores we use a popular metric from information retrieval. Recall at top $k$ predictions (ranked by model confidence scores) can give an intuitive indication if the model can retrieve the desired ground truth diseases. The evaluation is in line with prior work e.g. by \citet{cgl-lu-ijcai21, sherbet-lu-ieee21, chet-lu-aaai22}.

\subsubsection{Readmission}
See details in Sec. \ref{sec:experiments}. We define a readmission task at a horizon $h$ for a given patient history at time $t$ and a subsequent admission at time $t_{readm.}$ the target is defined as $y_{readm.@h} = (t_{readm.} - t) < h$. (Emergency) readmissions are clinically highly relevant as shown by nation-wide deployments of such systems in e.g. Scotland \citep{sparra-liley-medrxiv21}. We evaluate the performance using \emph{AuROC}.

\begin{align}
\label{eqn:detail-readmission-task}
    \textit{days}(v_{t+1} - v_{t}) < horizon &= \textit{model} \bigl( \mathcal{C}_{i, 0\ldots t}(*) \bigr) \\
    & \text{where}\,* \in \{{d, m, n}\} \nonumber
\end{align}

\subsubsection{Length of Stay}
We perform a length of stay prediction (\emph{LOS}) using a multi-class approach with 10 categories (\cite{pyhealth-yang-kdd23}. 0 for stays under 1 day, 1-7 for stays with the respective length in number of days, 8 for stays from 1 to two weeks, and 9 for stays over 2 weeks). To evaluate the imbalanced multi-class prediction task we use \emph{weighted-AuROC}.

\begin{align}
\label{eqn:detail-readmission-task}
    \textit{duration}(v_{t}) &= \textit{model} \bigl( \mathcal{C}_{i, 0\ldots t}(*) \bigr) \\
    & \text{where}\,* \in \{{d, m, n}\} \nonumber
\end{align} 
\subsection{Baselines}
\label{apd:baselines}

In this section, we provide a summary overview of the presented baselines and the key points of their architectures.

\subsubsection{CGL: Collaborative Graph Learning}
In this work, \citet{cgl-lu-ijcai21} propose a collaborative graph learning approach. They consider two graphs, one where patients and diseases are connected based on co-occurrence and one where only diseases are connected amongst each other based on the ICD ontology. GNN layers over the two edge sets and the shared set of nodes are run in an interleaved fashion (collaboratively). The computed embeddings for a certain disease are aggregated to represent patient visits and a sequence model performs task predictions.

\subsubsection{Chet: Context aware Health Event Prediction via Transition Functions}
The core contribution of this work by \citet{chet-lu-aaai22} is to consider a global disease graph, which connects diseases by co-occurrence and ontology relations, as well as a local graph (for each visit), which models the interactions of assigned disease codes within this specific visit. The architecture includes aggregation functions and sequence modeling to perform task-specific predictions.

\subsubsection{Sherbet: Self- Supervised Graph Learning With Hyperbolic Embedding for Temporal Health Event Prediction}
With \emph{Sherbet} \citep{sherbet-lu-ieee21} propose to encode the structure of a disease ontology in hyperbolic space. The hyperbolic embeddings for the respective diseases are used to pretrain (using a patient history reconstruction task) and fine-tune a sequence model architecture to perform task-specific predictions.

\subsubsection{MedPath: Augmenting Health Risk Prediction via Medical Knowledge Paths}
With \emph{MedPath} \citet{medpath-ye-www21} propose to enhance the performance of existing EHR representation learning architectures by incorporating a \emph{personalized} graph extracted using knowledge from \emph{Semantic MEDLINE} \citep{semmed-rindflesch-isu11}. The extracted graph is dataset and task-specific and can improve the performance of the backbone architecture. We transformed our data to adapt to their published pipeline and performed the heart failure prediction task using their implementations. We use \emph{HiTANet} \citep{hitanet-luo-kdd20} as the backbone architecture, because it performed the best on the validation set in our hyperparameter search.

\subsubsection{G-BERT: Pre-training of Graph Augmented Transformers for Medication Recommendation}
\citet{gbert-shang-ijcai19} show performance improvements on a medication recommendation task by pretraining disease and medication code embeddings using GNNs over two ontologies. The pretraining objective is a reconstruction task of observed codes during a patient visit and borrows ideas from masked language modeling. The pretrained architecture includes a Transformer-based encoder, which outputs a \texttt{CLS} encoding for each patient visit. The proposed downstream architecture performs a pooling scheme over patient histories and recommends medications for a current patient visit given the patient's history and the current diagnosis of diseases.

\subsubsection{Embedding Matrix}
\label{app:embedding-matrix}
In \tableref{tab:concept-embedding-ablation} we show a concept embedding ablation using an \emph{Embedding Matrix}. This refers to a matrix of trainable parameters $\mathbf{E} \in \mathbb{R}^{|\mathcal{C}| \times k}$ where $|C|$ is the total number of considered medical concepts and $k$ the embedding dimension. The embedding matrix replaces the \emph{Concept Embedding} (Sec. \ref{sec:concept-embedding}) module and is pretrained and fine-tuned using the same procedure.

\subsubsection{Concept Embeddings using Node2Vec}
In \tableref{tab:concept-embedding-ablation} we show a concept embedding ablation using Node2Vec \citep{node2vec-grover-kdd16}. We consider our extracted complex UMLS-based knowledge graph and perform Node2Vec-style pretraining to obtain embeddings for each concept in our knowledge graph.
We then initialize an embedding matrix (which is used to retrieve concept embeddings by index lookup) and use it to replace our proposed GNN-based concept embeddings. To ensure fair comparison we then perform the same reconstruction pretraining as our proposed approach \emph{MMUGL} to ensure the parameters of the Visit Encoder (Sec. \ref{sec:visit-encoding}) module are well pretrained too. Similarly, we apply the same pipeline as for our approach during fine-tuning for downstream tasks.

\subsubsection{Concept Embeddings using Cui2Vec}
\emph{Cui2Vec} as introduced by \citet{cui2vec-beam-biocomp20} is a collection of pretrained medical concept embeddings mapped to the space of UMLS. Their training optimizes a \emph{Word2Vec}~\citep{w2v-mikolov-arxiv} style objective over a large-scale corpus (60 million patient records, 20 million clinical notes, and 1.7 million full-text biomedical journal articles). We use the \emph{Cui2Vec} embeddings to initialize a lookup matrix from which concept embeddings are retrieved by index and replace our GNN-based concept embeddings. To ensure fair comparison we apply the same pretraining (reconstruction) and fine-tuning procedure to obtain downstream task performance results. 
\section{Training and Architecture Ablations}

\subsection{Clinical Reports Performance Contribution}
\label{apd:clinical-reports-performance}

In this section, we would like to clarify our findings about why the additional modality of extracted concepts from unstructured text (i.e. clinical reports) cannot yield a performance improvement in all cases.

Overall, the billing codes represent an aggregate of information for an entire patient's visit to the hospital and the labels are defined based on them. Thus, the billing codes (ICD, ATC codes) are the strongest signal for our predictions. The additional medical concepts from clinical reports can help in two ways. First, they can help to deal with missing or noisy information from the billing codes (see also Appendix~\ref{apd:missing-information}). Second, they can help the model to make more fine-grained predictions due to the higher level of detail.

\paragraph{Heart Failure} Here the additional concepts from clinical reports do seem to help, but in most cases only marginally. We hypothesize this is due to the fact, that we are only performing a binary prediction and the finer details of the clinical reports cannot yield enough additional information in most cases to significantly improve our predictions.

\paragraph{Diagnosis} This is a very complex classification task and here we see the strongest improvement after adding the concepts extracted from clinical reports. For this task, we can benefit from the higher level of detail present in the clinical reports compared to the billing codes.

\paragraph{Medication Recommendation} On this task the strongest signal comes from the current set of diseases. The additional concepts from clinical reports are only present in the representation of the patient's history, where we do not seem to benefit from the more detailed content of the clinical reports. To avoid information leakage we cannot directly use all concepts from all reports of the current visit when performing the medication recommendation. To accommodate for this we would have to adapt the task to a within-visit online medication recommendation; predicting medication based on the patient's global (past hospital visits) as well as local (past time within current visit) history. This would enable the inclusion of already accumulated clinical reports in the local (current visit) context. 
\subsection{Ablation: SapBERT}
\label{apd:ablation-sapbert}
\label{app:sapbert}
We ablate the use of \emph{SapBERT}~\citep{sapbert-liu-naacl21} compared to training randomly initialized node embeddings. \emph{SapBERT} performs better in pretraining (the selection criteria), where we see an increase from $49.38\pm0.49$ to $61.77\pm0.44$ \emph{AuPRC}. The improvement carries over to the downstream performance, where for the diagnosis prediction we see an improvement of $25.46\pm0.50$ to $26.19\pm0.30$ in the \emph{F1 (inflated)} score.
 
\subsection{Pretraining Ablation}
\label{apd:pretraining-ablation}

\begin{table}[tb]
    \centering
    \caption{Ablation for pretraining, data splits, and target code sets of CGL~\citep{cgl-lu-ijcai21} on the \emph{Diagnosis} task; best in bold. $\mathcal{C}(n)$ refers to med. concepts from text. Details in Section~\ref{sec:exp-pretraining-ablation}.}
    \setlength{\tabcolsep}{4pt}
    \begin{tabular}{l l l}
    \toprule
    \bf Method & \bf \textit{w-F1} & \bf \textit{w-F1} \\ & random init. & pretrained \\
    \midrule

    ICD/ATC                     & 13.68$\pm$0.62            & 24.18$\pm$0.24 \\ ICD/ATC-CO                  & 16.69$\pm$0.39            & 24.11$\pm$0.24 \\ \emph{MMUGL w/o $\mathcal{C}(n)$}        & \textbf{20.78$\pm$0.21}   & 24.24$\pm$0.34 \\ \emph{MMUGL}                 & 20.18$\pm$0.52            & \textbf{24.69$\pm$0.21} \\

    \bottomrule
    \end{tabular}
    \label{tab:pretraining-ablation}
\end{table} 
\label{sec:exp-pretraining-ablation}

In \tableref{tab:baseline-comparison-diag} we can see, that prior work including pretraining schemes performs much stronger than the ones that don't.
In \tableref{tab:pretraining-ablation} we perform an ablation w.r.t.~pretraining different concept embeddings and report performance on the \emph{Diagnosis} task (Sec.~\ref{sec:diagnosis-task}) on pretrained and on randomly initialized networks. We note, the more structure bias we provide, the better the performance without pretraining.
In Appendix~\ref{apd:sum-loss} and~\ref{apd:recon-loss} we present further results on exploring modifications to the pretraining loss function. 
\subsection{Sum Aggregation Loss}
\label{apd:sum-loss}
\begin{figure}[tb]
    \centering
    \includegraphics[width=\linewidth]{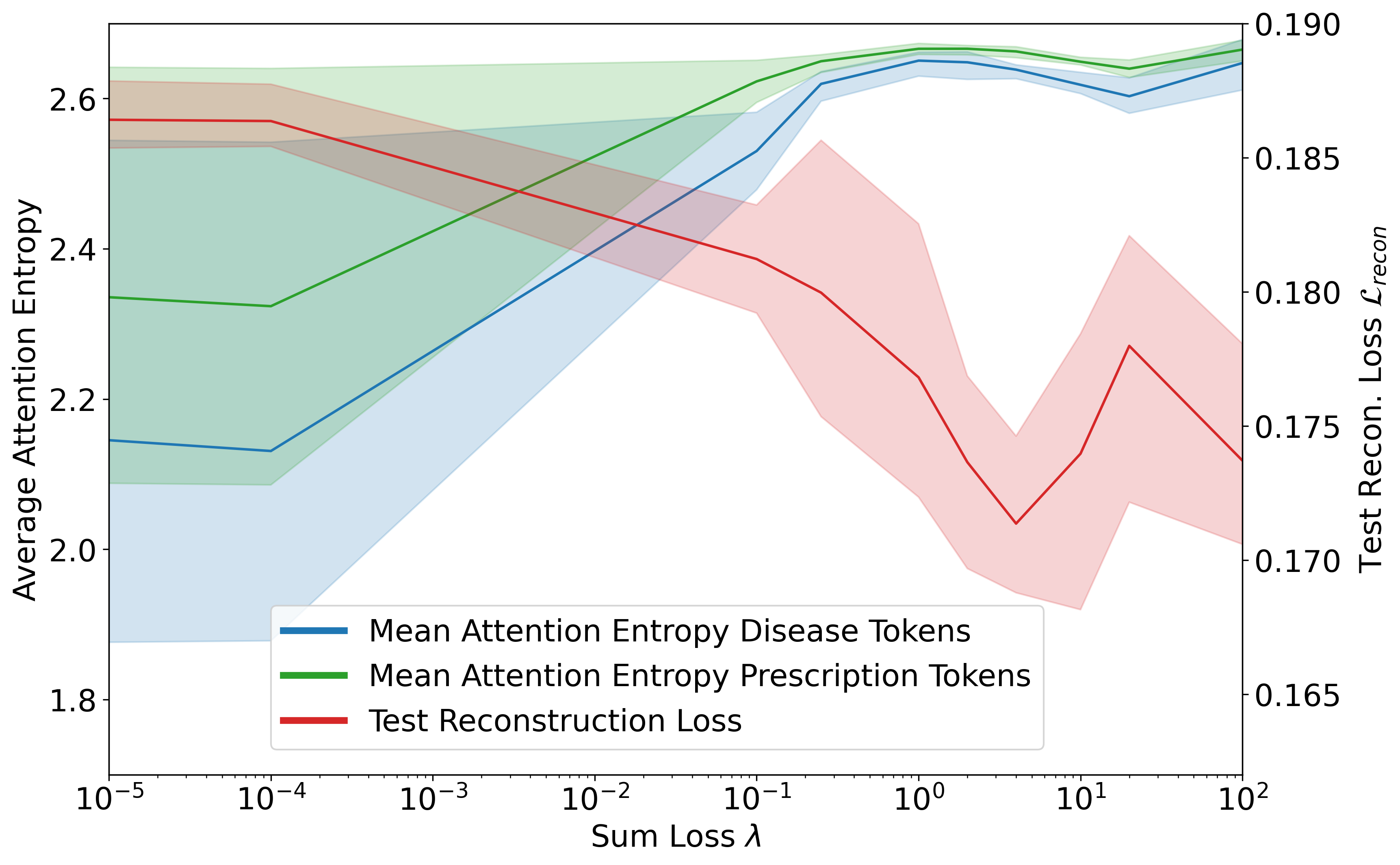}
    \caption{Behavior of average \texttt{CLS} attention entropy and pretraining test reconstruction loss $\mathcal{L}_{recon}$ w.r.t.~$\lambda$ parameter of $\mathcal{L}_{sum}$ (Eqn.~\ref{eqn:sum-pretraining-loss}). Details in Sec.~\ref{apd:sum-loss}.}
    \label{fig:pretraining-sum-loss-entropy}
\end{figure}

In \figureref{fig:pretraining-sum-loss-entropy} we perform an ablation w.r.t.~to the hyperparameter $\lambda$ controlling the contribution of $\mathcal{L}_{sum}$ (Eqn.~\ref{eqn:sum-pretraining-loss}) to the total pretraining loss.
\citet{analyzing-transformers-entropy-vig-acl19} have computed the entropy of the distribution induced by the attention mechanism to analyze Transformer behavior. Similarly, we show the average (test set) entropy of the distribution induced by attention from the \texttt{CLS} token to all the other tokens. For larger $\lambda$ the entropy increases, hence the distribution is more dispersed, and we can see an improvement in pretraining performance (shown by the test set reconstruction loss $\mathcal{L}_{recon}$ corresponding to improved test log-likelihood of our model). The idea is, that a more dispersed distribution is a better aggregator and generalizes better to rare diseases, which might otherwise be overlooked by a pointy (overfitted) attention distribution.
In Appendix~\ref{apd:sum-loss} and \ref{apd:pretraining-ablation} we provide further experimental results ablating pretraining and the different loss terms.

We provide further empirical evidence for the contribution of the additional loss term introduced in Eqn.~\ref{eqn:sum-pretraining-loss} in \tableref{tab:sum-loss-diagnosis}.

\begin{table*}[htb]
\floatconts
    {tab:sum-loss-diagnosis}
    {
        \caption{
            Ablation for pretraining considering data splits and target code sets of CGL~\citep{cgl-lu-ijcai21} on the \emph{Diagnosis} task.
            We show results on the downstream network without pretraining, with pretraining only using $\mathcal{L}_{recon}$ Eqn. \ref{eqn:auto-encoder-loss},
            and with pretraining including $\mathcal{L}_{sum}$ Eqn. \ref{eqn:sum-pretraining-loss}.
            $w_{\bullet, m}$ refers to Eqn. \ref{eqn:auto-encoder-loss-weighted}. Bold font highlights the best in each row. Details in Appendix~\ref{apd:sum-loss}.
        }
    }
    {
        \begin{tabular}{l l l l}
        \toprule
        \bf Method & \bf \textit{w-F1} & \bf \textit{w-F1} & \bf \textit{w-F1} \\ 
          & random init. & pretrain w/o $\mathcal{L}_{sum}$ & pretrain w. $\mathcal{L}_{sum}$ \\
        \midrule
    
        ICD/ATC                     & 13.68$\pm$0.62            & 23.84$\pm$0.39    & \textbf{24.18$\pm$0.24} \\
        ICD/ATC-CO                  & 16.69$\pm$0.39            & 23.30$\pm$0.79    & \textbf{24.11$\pm$0.24} \\
        \emph{MMUGL w/o Text}        & 20.78$\pm$0.21            & 24.05$\pm$0.34    & \textbf{24.24$\pm$0.34} \\
        \emph{MMUGL}                 & 20.18$\pm$0.52            & 24.62$\pm$0.22    & \textbf{24.69$\pm$0.21} \\
        \emph{MMUGL}, $w_{\bullet, m} = 0$ & \multicolumn{1}{c}{-}                   & 24.42$\pm$0.86    & \textbf{25.39$\pm$0.09} \\

        \bottomrule
        \end{tabular}
    }
\end{table*}

\tableref{tab:sum-loss-diagnosis} shows results on the \emph{Diagnosis} downstream task across different \emph{Concept Embedding} implementations
and with different pretraining regimes. We show results without pretraining, pretraining on only the default reconstruction loss $\mathcal{L}_{recon}$ (Eqn.~\ref{eqn:auto-encoder-loss}) and including the additional introduced loss term $\mathcal{L}_{sum}$ (Eqn.~\ref{eqn:sum-pretraining-loss}).

We can see that the additional loss component $\mathcal{L}_{sum}$ during pretraining contributes to better pretrained representations as across different
downstream models we can see either at least the same performance or increased performance. This difference is especially notable and important for the
best-performing model implementation \emph{MMUGL}, where $w_{\bullet, m} = 0$ (Eqn.~\ref{eqn:auto-encoder-loss-weighted}, pretraining focused on recovering diseases only).
We hypothesize, that without the additional loss regularization, we experience stronger overfitting to the training distribution during pretraining, as we have more data
available (given that \emph{MMUGL} includes additional rich information coming from medical concepts in clinical reports) and we have reduced the task complexity (as we set $w_{\bullet, m} = 0$ in the pretraining loss, Eqn. \ref{eqn:auto-encoder-loss-weighted}). We also observe a tendency to more consistent results under pretraining
including the $\mathcal{L}_{sum}$ loss component as standard deviations tend to be lower.

This stays consistent also on a further task e.g. \emph{Heart Failure}. For \emph{MMUGL} with $w_{\bullet, m} = 0$ including $\mathcal{L}_{sum}$ in pretraining
we observe a downstream heart failure prediction performance (on the CGL~\cite{cgl-lu-ijcai21} patient split) of $87.60\pm0.40$ where this drops to $86.93\pm0.13$
if we pretrain without $\mathcal{L}_{sum}$.

\subsection{Reconstruction Loss}
\label{apd:recon-loss}

\begin{table*}[htb]
\floatconts
  {tab:recon-loss-ablation}
  {\caption{
        Ablation for the pretraining reconstruction loss $\mathcal{L}_{recon}$ (Eqn.~\ref{eqn:auto-encoder-loss})
        considering data splits and target code sets of CGL~\citep{cgl-lu-ijcai21} on the \emph{Diagnosis} task.
        The weights $w_{\bullet, *}$ refer to Eqn.~\ref{eqn:auto-encoder-loss-weighted}. The results shown are
        computed on the model variant \emph{MMUGL w/o Text}. Best in bold. Details in Appendix~\ref{apd:recon-loss}.
    }}
  {
    \begin{tabular}{c c c c r r r}
    \toprule
$w_{d, d}$ & $w_{m, d}$ & $w_{m, m}$ & $w_{d, m}$ & \textit{w-F1} & \textit{R@20} & \textit{R@40} \\
    \midrule

1 & 0 & 0 & 0 &     23.18$\pm$0.20 & 39.30$\pm$0.42 & 51.03$\pm$0.26	\\
    1 & 1 & 0 & 0 &     \textbf{24.46$\pm$0.18} & \textbf{40.76$\pm$0.10} & 52.14$\pm$0.31    \\
    1 & 1 & 1 & 0 &     24.15$\pm$0.56 & 40.39$\pm$0.07 & 51.61$\pm$0.17	\\
    1 & 1 & 1 & 1 &     24.24$\pm$0.26 & 40.49$\pm$0.02 & \textbf{52.15$\pm$0.23}    \\

    \bottomrule
    \end{tabular}
  }
\end{table*}

\begin{table*}[htb]
\floatconts
  {tab:recon-loss-ablation-med}
  {\caption{
        Ablation for the pretraining reconstruction loss $\mathcal{L}_{recon}$ (Eqn.~\ref{eqn:auto-encoder-loss})
        considering data splits and target code sets of G-BERT~\citep{gbert-shang-ijcai19} on the \emph{Medication Recommendation} task.
        The weights $w_{\bullet, *}$ refer to Eqn.~\ref{eqn:auto-encoder-loss-weighted}. The results shown are
        computed on the model variant \emph{MMUGL w/o Text}. Best in bold. Details in Appendix~\ref{apd:recon-loss}.
    }}
  {
    \begin{tabular}{c c c c r r}
    \toprule
$w_{d, d}$ & $w_{m, d}$ & $w_{m, m}$ & $w_{d, m}$ & \textit{AuPRC} & \textit{F1} \\
    \midrule

0 & 0 & 1 & 0 &     68.00$\pm$0.24 & 59.97$\pm$0.43	\\
    0 & 0 & 1 & 1 &     \textbf{72.44$\pm$0.07} & \textbf{63.58$\pm$0.15} \\
    1 & 0 & 1 & 1 &     72.37$\pm$0.11 & 63.51$\pm$0.11	\\
    1 & 1 & 1 & 1 &     72.09$\pm$0.02 & 63.34$\pm$0.04 \\

    \bottomrule
    \end{tabular}
  }
\end{table*}

In \tableref{tab:recon-loss-ablation,tab:recon-loss-ablation-med} we perform an ablation with respect to the different weights
in the weighted version of the pretraining reconstruction loss $\mathcal{L}_{recon}$ (Eqn.~\ref{eqn:auto-encoder-loss-weighted}).
The base version as introduced by~\citet{gbert-shang-ijcai19} considers all weights $w_{\bullet, *} = 1$. This is flexible
in the sense that it does not enforce a bias towards encoding information relevant for disease or medication predictions.
However, by weighting (or fully disabling) the different terms, we can tailor our pretraining to different downstream scenarios.
Please also note, that the following experiments have been performed without the additional loss component $\mathcal{L}_{sum}$ (Eqn.~\ref{eqn:sum-pretraining-loss}) to focus purely on the effects within
the reconstruction loss term $\mathcal{L}_{recon}$ (Eqn.~\ref{eqn:auto-encoder-loss},~\ref{eqn:auto-encoder-loss-weighted}).

\paragraph{Downstream Diagnosis} \tableref{tab:recon-loss-ablation} shows this effect on the downstream \emph{Diagnosis} task. We can see that while
having all loss terms active yields strong performance, in the case of a diagnosis prediction it is beneficial to
only pretrain on loss terms that are predictive for diseases i.e. $w_{\bullet, d} = 1 \wedge w_{\bullet, m} = 0$.
This is further supported by results shown in \tableref{tab:baseline-comparison-diag} and \tableref{tab:concept-embedding-ablation},
where results on the full \emph{MMUGL} model (including medical concepts from clinical reports) improve by pretraining with $w_{\bullet, m} = 0$.

\paragraph{Downstream Medication Recommendation} \tableref{tab:recon-loss-ablation-med} shows the exact same behaviour when performing downstream \emph{medication recommendation}. The best performance is achieved by only considering loss terms towards predicting the modality relevant for the downstream prediction task. We can conclude, that cross-modality pretraining is beneficial to learn embeddings that can be useful for a yet unspecified downstream application. However, if the nature of
the target modality of the downstream task is known and the cost of pretraining affordable, we can achieve
better performance by adapting the pretraining to the downstream scenario. 
\subsection{Knowledge Graph Scaling}
\label{apd:kg-scaling}
We perform a small set of experiments to investigate the effect of reducing the size of our knowledge graph. We reduce the graph to 50\% of its original size (App.~\ref{apd:data}) by dropping concepts with lower frequencies first.

We observe for diagnosis $26.41\pm0.12$ (full size graph: $26.40\pm0.02$, Table~\ref{tab:baseline-comparison-diag}). However, for readmission at a 1 year horizon we observe $73.11\pm1.87$ (full size graph: $73.26\pm0.18$, Table~\ref{tab:concept-embedding-ablation}), which is again similar but with much higher variance (over 3 random seeds). For readmission at a 2 year horizon we observe $77.27\pm1.36$ (full size graph: $78.01\pm0.91$, Figure~\ref{fig:readmission-multiple-horizons}). Some task seem to remain stable at smaller scale, whereas others show slight decrease in performance or higher variance. 

\section{Clinical Report Concept Category Distribution}
\label{apd:text-concept-categories}
\begin{figure*}[htb!]
\floatconts
  {fig:apd-text-attention-lin-log-comparison}
  {\caption{Medical concepts from clinical reports (Sec.~\ref{sec:nlp-pipeline}) ranked by the visit encoders attention scores. We extract the top-8 tokens, group them by note type and compare the distribution of all tokens in the dataset with the ranked top-k distribution. Details in Section~\ref{sec:interpretability-analysis-text-types} and Appendix~\ref{apd:text-concept-categories}.}}
  {\subfigure[Logarithmic y-axis]{\label{fig:apd-text-attention-log}\includegraphics[width=.45\linewidth]{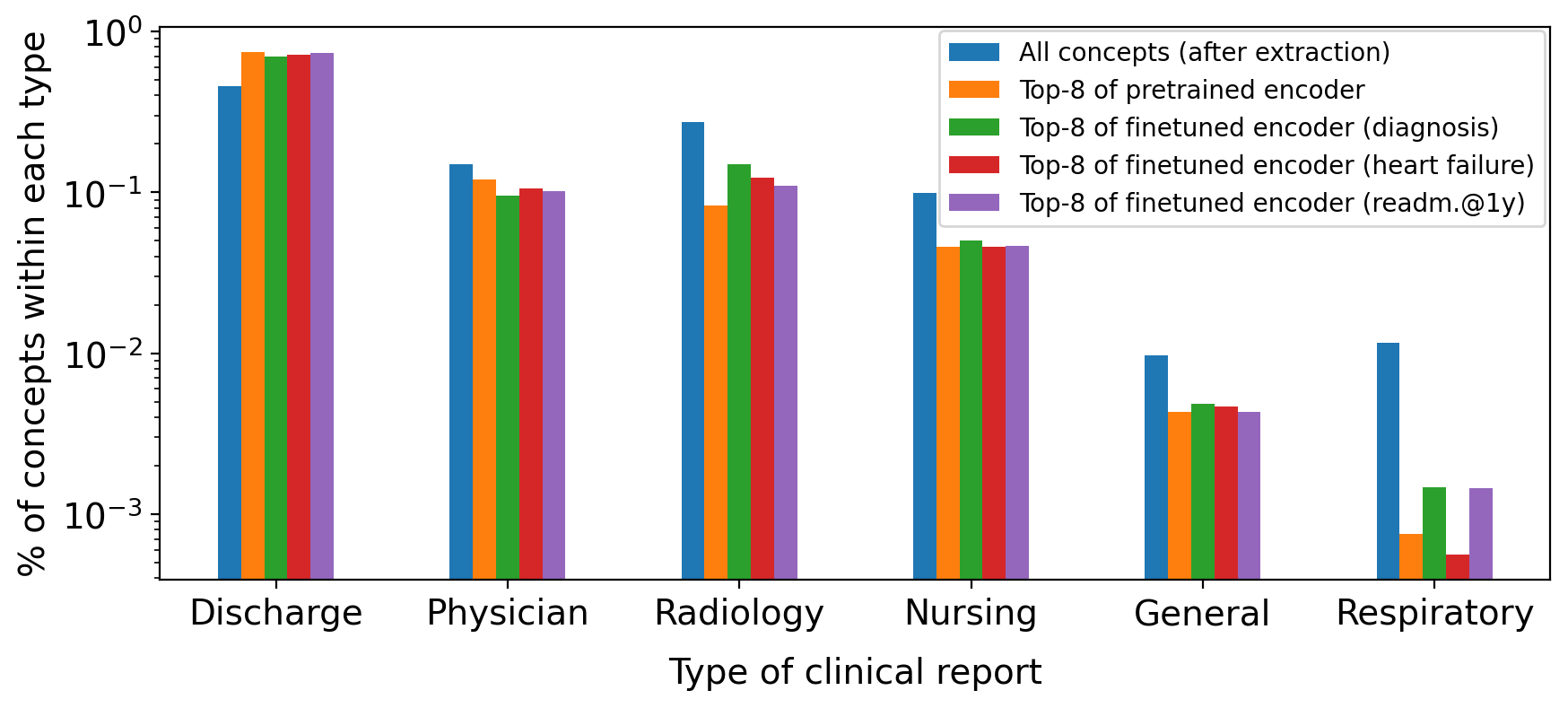}}\qquad
    \subfigure[Linear y-axis]{\label{fig:apd-text-attention-lin}\includegraphics[width=.45\linewidth]{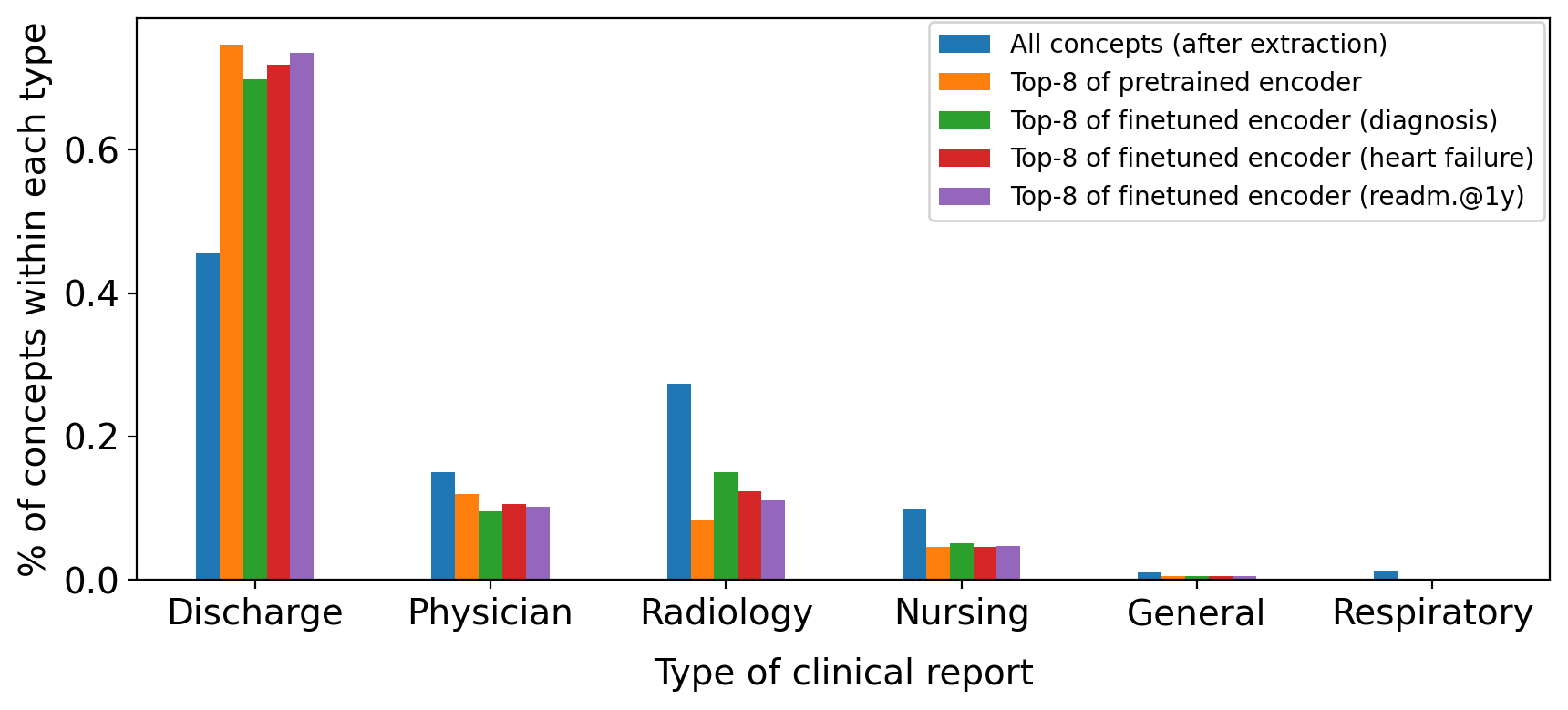}}
  }
\end{figure*}

We can use the attention mechanism to interpret the results on a patient level to rank diagnosis and medications, as well as general medical concepts from clinical reports w.r.t.~their importance for the prediction. An example of such an analysis is in Appendix \ref{apd:single-patient-inter}.

We can also perform various dataset global analyses. We analyze the distribution of medical concepts extracted from clinical reports w.r.t.~MIMIC-III report type and present the results in \figureref{fig:apd-text-attention-log}; please mind the logarithmic y-scale (Appendix~\ref{apd:text-concept-categories} also shows a linear scale). After pretraining, we can see a very strong shift from the dataset's type distribution toward discharge summaries. This is sensible given the pretraining task is an auto-encoder, essentially training for \emph{summarizing} the visit. By fine-tuning for specific tasks we can see slight shifts towards more specific report types, which can help provide more detailed insights for a given task; note for example how the focus in the respiratory category increases as we fine-tune for a general diagnosis, but decreases below the pretraining level for a heart failure prediction.

The plot with logarithmic scale in Figure \ref{fig:apd-text-attention-log} is better suited to highlight the fine details and changes in categories such as \emph{Respiratory} or \emph{Radiology}. The linear scale in \figureref{fig:apd-text-attention-lin} shows the strong changes caused by the pretraining (compared to the actual token distribution per category) in e.g. the \emph{discharge summary} type of reports.

One might notice a particularly large drop in tokens from reports of the respiratory type. First we would like to highlight that \figureref{fig:apd-text-attention-log} uses a logarithmic y-axis and thus the absolute number of tokens found in the respective report type is comparatively low. Still, we can observe a change over one order of magnitude. This can be explained by looking more in-depth at the reports of this specific type.

In MIMIC-III the clinical reports of type \emph{Respiratory} are mostly highly structured status reports assessing a patient's state w.r.t. the respiratory system. Being a structured report, there is a large set medical concepts matched, which correspond to the field names of the structured report to be filled with patient information and further most of the provided assessments in the form do not vary much across patients. As such, many of the extracted medical concepts from these reports are not discriminative across patients and thus we observe a drop in attention to the tokens extracted from these reports after training the model.

\section{Heart Failure Performance Disentanglement}
\label{apd:hf-performance-ablation}

Due to the chronic nature of heart failure, we disentangle the performance on the test set with a fixed model for patients with and without reported histories of heart failure (the target codes have appeared in the patient history). The results are shown in Table~\ref{tab:hf-history-ablation}.

\begin{table*}[tb]
    \centering
    \caption{Performance disentanglement for the heart failure prediction task on patients with and without reported history of heart failure. \textit{AuROC} with text shows the full model performance, and \textit{AuROC} w/o text the performance when all concepts extracted from clinical reports are masked.}
    \setlength{\tabcolsep}{4pt}
    \begin{tabular}{l r r r r r}
    \toprule
    Patient Set & \# Patients & Prevalence & \multicolumn{3}{c}{\textit{F1}} \\
     &  &  & w/o text & w. text & $\Delta$ text \\
    \midrule

All & 1221 & 37\% & 73.29 & 75.08  & +1.79 \\
    W. HF History & 425 & 74\% & 85.52 & 86.28 & +0.76  \\
    W/o HF History & 796 & 17\% & 44.02 & 47.68 & +3.66 \\

    \bottomrule
    \end{tabular}
    \label{tab:hf-history-ablation}
\end{table*}

The model is naturally performing much better on the subset of patients with a reported history of heart failure and can exploit the chronic nature of the disease. However, we note that with our proposed multi-modal approach we see a notable performance improvement on the hard cases of patients without a reported history of heart failure. We conclude, using clinical report concepts backed by a knowledge graph, not just billing codes, aids in understanding disease progressions.

\section{Single Patient Interpretability}
\label{apd:single-patient-inter}

\begin{figure*}[htb]
\floatconts
  {fig:apd-single-patient}
  {
    \caption{Attention score analysis for a single visit of a single patient (MIMIC-III~\citep{mimic-iii-johnson-nature16} dataset). We extract the scores for the different modalities from our visit encoder $g_{\psi}(v)$. Details in Appendix~\ref{apd:single-patient-inter}.}
  }
  {
    \subfigure[Overview of attention scores for different modalities. We show categories for ICD and ATC codes, aggregated scores for each clinical report within the visit and an aggregate for each category of reports.]{\label{fig:apd-patient-overview}\includegraphics[width=.75\linewidth]{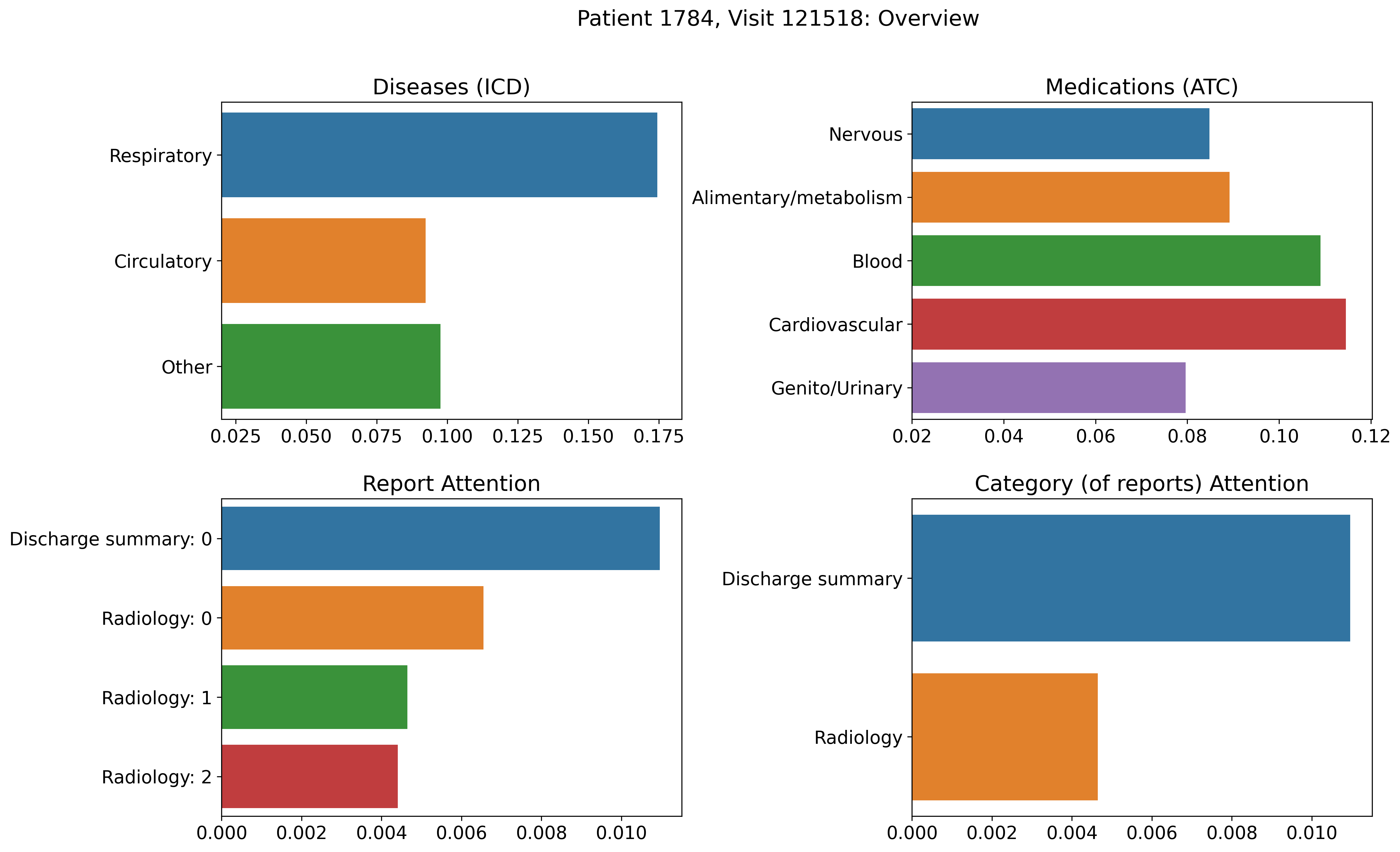}}\vspace{1em}
    \subfigure[We show the 10 top ranked UMLS concepts within the two notes with highest aggregated attention score (see Subfigure \ref{fig:apd-patient-overview}).]{\label{fig:apd-patient-top-reports}\includegraphics[width=.75\linewidth]{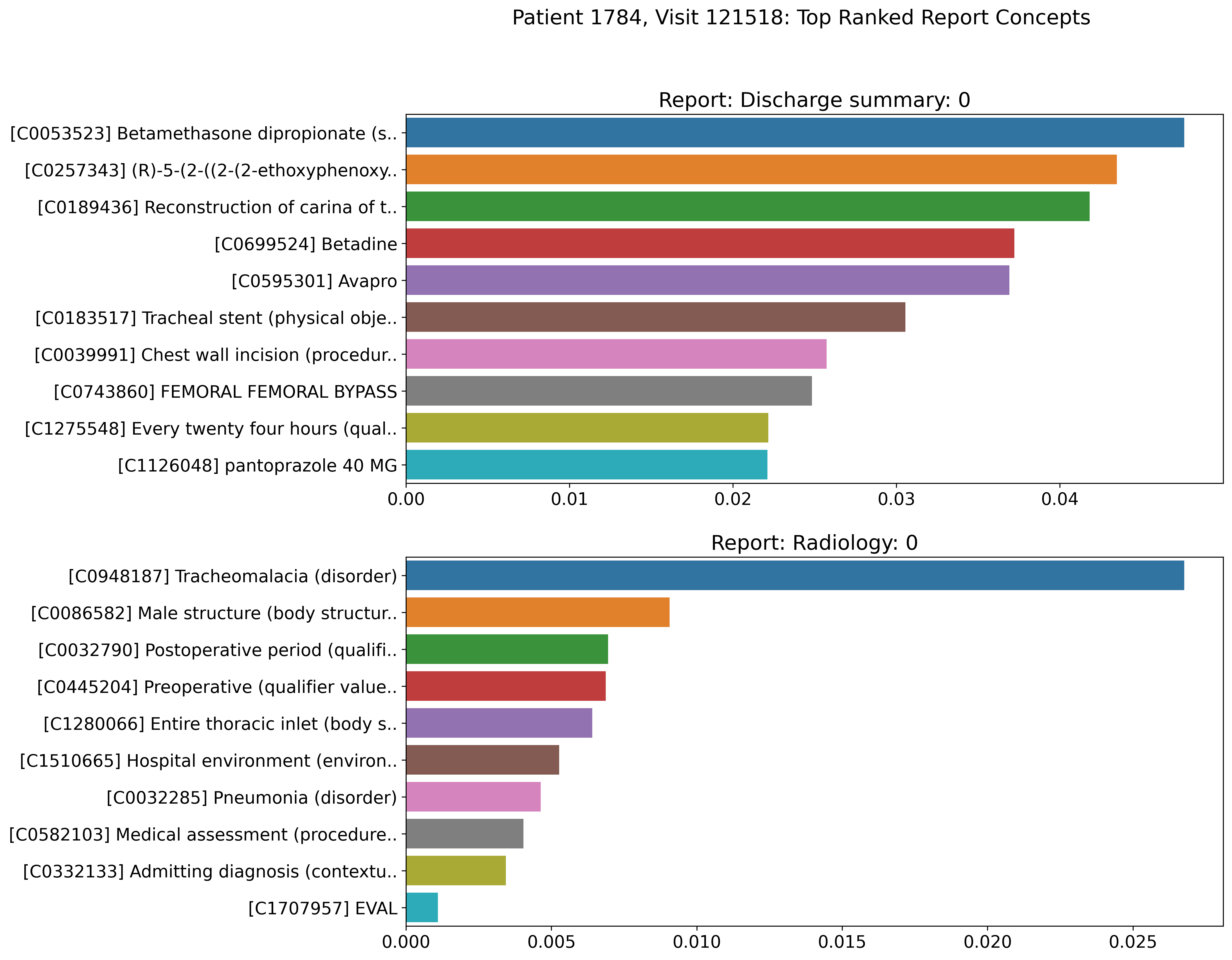}}
  }
\end{figure*}

In \figureref{fig:apd-single-patient} we present various ways how attention scores of our visit encoder (Sec.~\ref{sec:visit-encoding}) can be used to provide interpretability of our predictions. We provide an example score analysis of visit $121518$ by patient $1784$ in the MIMIC-III~\citep{mimic-iii-johnson-nature16} dataset. The patient was assigned the following set of codes:

\begin{itemize}
    \item \emph{ICD}: 519.1, 496.0, 414.01, 401.9, 443.9, V45.82
    \item \emph{ATC4}: N05CD, A02BC, B01AB, A06AD, C07AB, B05CX, G04CA, A07EA
\end{itemize}

The scores can be used to highlight the most relevant diseases and medications (\figureref{fig:apd-patient-overview}). By grouping scores of individual codes and computing an aggregate for each group (e.g. $90th$-percentile of scores) we can highlight the most relevant disease and medication categories for this patient at the given visit.

We can further extract which of the reports collected during the entire visit contain the most predictive identifiers by computing an aggregated score over the scores of all the matched concepts within each report (\figureref{fig:apd-patient-overview}).

In (\figureref{fig:apd-patient-top-reports}) we then highlight the concepts within the two highest-ranked reports with the largest attention scores.

 We can see that the scores are consistent across different modalities, considering for example the high score given to the Respiratory category for the disease (ICD) codes (\figureref{fig:apd-patient-overview}), as well as high scores for concepts found in clinical reports (e.g. \texttt{C0948187} (Tracheomalacia) in \texttt{Note: Radiology~0} or \texttt{C0189436} (Carinal reconstruction) in \texttt{Note: Discharge Summary:~0}; \figureref{fig:apd-patient-top-reports}) related to respiratory conditions. We can conclude that for this sample the unified concept latent space promotes consistency across modalities and can improve interpretability.

\section{Robustness w.r.t. missing information}
\label{apd:missing-information}
In \figureref{fig:masking-progression} we show the results of an experiment, where we progressively mask a larger percentage of input tokens of different modalities. This is done by replacing the respective token identifier with the \texttt{MASK} token used during masked language modeling style pretraining \citep{gbert-shang-ijcai19, bert-devlin-naacl19}.

Tokens can either be masked randomly or we sort them with respect to the attention score assigned to them in the visit encoder. The y-axis shows the pretraining performance w.r.t to Eqn.~\ref{eqn:auto-encoder-loss}; decoding to any of the two modalities (diseases, medications) from the visit representation of either.

The results show, that although the auto-encoding objective is only formulated w.r.t. the disease and medications tokens, the additional text information can successfully prevent stronger decay in performance and help impute the missing or incorrect information.

We can further see that masking tokens according to their attention scores results in a faster overall decrease in performance, highlighting the benefits of using an attention-based encoder, that can focus on relevant medical concepts when encoding a patient's current state.

\begin{figure*}[htb]
\floatconts
  {fig:masking-progression}
  {
      \caption{We analyze the performance decay caused by progressively masking a larger percentage of tokens of certain input types. We observe, that the additional information provided by text notes can have a large impact on performance, and can help prevent stronger decays in performance under missing information in the discrete token identifiers (ICD and ATC codes). Details in Appendix~\ref{apd:missing-information}.}
  }
  {\includegraphics[width=.8\linewidth]{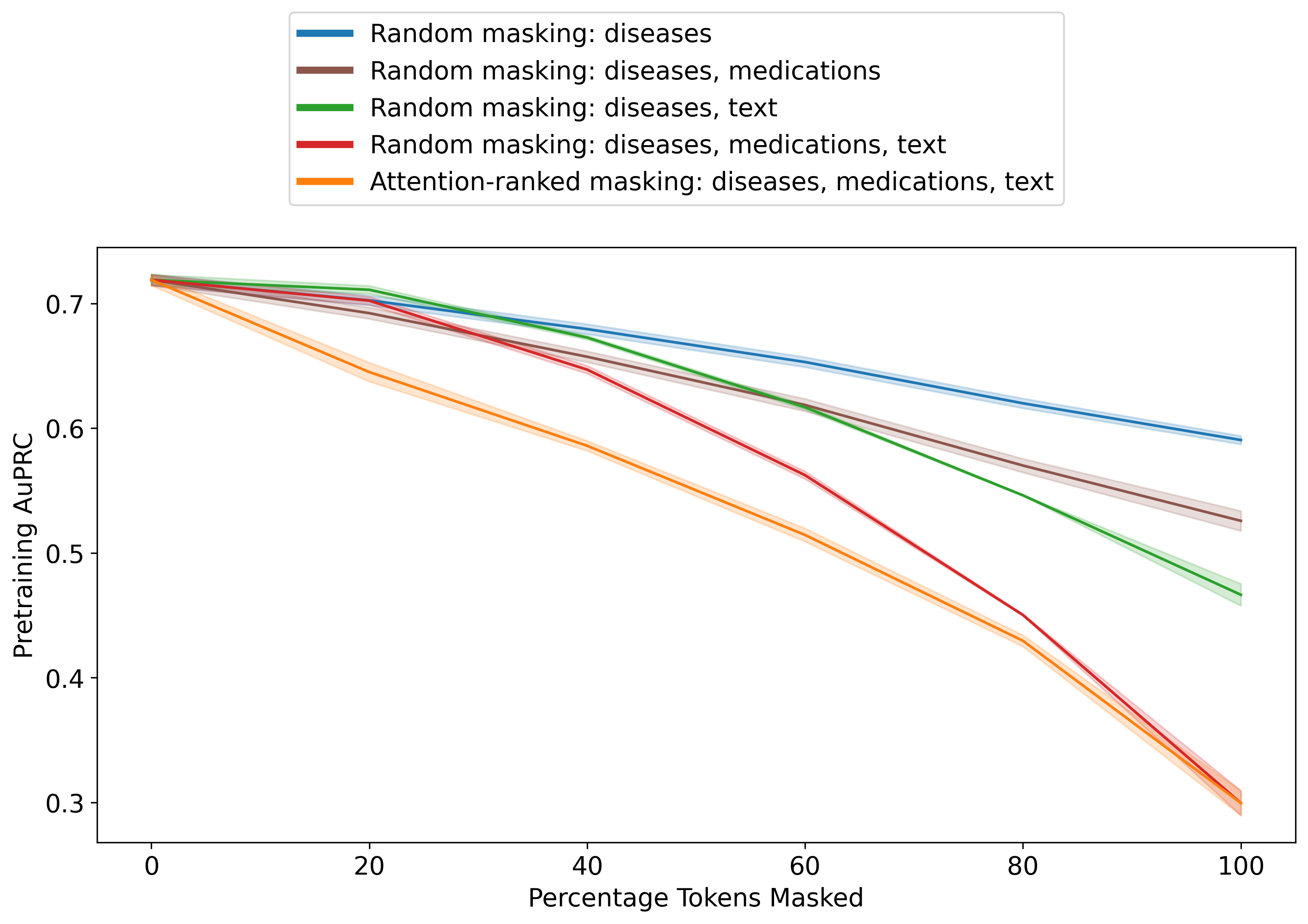}}
\end{figure*} 
\end{document}